\documentclass[10pt,twocolumn,letterpaper]{article}

\usepackage{cvpr}              % To produce the CAMERA-READY version
\usepackage{graphicx}
\usepackage{amsmath}
\usepackage{amssymb}
\usepackage{booktabs}
\usepackage[accsupp]{axessibility}  % Improves PDF readability for those with disabilities.

\usepackage[pagebackref,breaklinks,colorlinks]{hyperref}

\usepackage[capitalize]{cleveref}
\crefname{section}{Sec.}{Secs.}
\Crefname{section}{Section}{Sections}
\Crefname{table}{Table}{Tables}
\crefname{table}{Tab.}{Tabs.}

\usepackage{multirow}
\usepackage{tabu,booktabs}
\usepackage[dvipsnames]{xcolor}

\makeatletter
\newcommand*\bigcdot{\mathpalette\bigcdot@{.5}}
\newcommand*\bigcdot@[2]{\mathbin{\vcenter{\hbox{\scalebox{#2}{$\m@th#1\bullet$}}}}}
\makeatother

\makeatletter
\newcommand*\smaller{\@setfontsize\smaller{8}{10}}
\newcommand*\smallest{\@setfontsize\smallest{6.5}{8}}

\def\cvprPaperID{6100} % *** Enter the CVPR Paper ID here
\def\confName{CVPR}
\def\confYear{2023}

\begin{document}

\title{Movies2Scenes: Using Movie Metadata to Learn Scene Representation}
%Learning Scene Representations Using Movie Metadata 
\author{Shixing Chen \quad Chun-Hao Liu \quad Xiang Hao \quad Xiaohan Nie \quad Maxim Arap \quad Raffay Hamid\\
Amazon Prime Video\\
{\tt\small \{shixic, chunhaol, xianghao, nxiaohan, maxarap, raffay\}@amazon.com}
%\and
}
\maketitle

\vspace{-0.0cm}
\begin{abstract}

\vspace{-0.3cm}\noindent Understanding scenes in movies is crucial for a variety of applications such as video moderation, search, and recommendation. However, labeling individual scenes is a time-consuming process. In contrast, movie level metadata (\textit{e.g.}, genre, synopsis, \textit{etc.}) regularly gets produced as part of the film production process, and is therefore significantly more commonly available. In this work, we propose a novel contrastive learning approach that uses movie metadata to learn a general-purpose scene representation. Specifically, we use movie metadata to define a measure of movie similarity, and use it during contrastive learning to limit our search for positive scene-pairs to only the movies that are considered similar to each other. Our learned scene representation consistently outperforms existing state-of-the-art methods on a diverse set of tasks evaluated using multiple benchmark datasets. Notably, our learned representation offers an average improvement of $7.9$\% on the seven classification tasks and $9.7$\% improvement on the two regression tasks in LVU dataset. Furthermore, using a newly collected movie dataset, we present comparative results of our scene representation on a set of video moderation tasks to demonstrate its generalizability on previously less explored tasks.

\end{abstract}
\vspace{-0.0cm}
\vspace{-0.5cm}
\section{Introduction}
\label{intro}

\noindent Automatic understanding of movie scenes is a challenging problem~\cite{Wu_2021_CVPR}~\cite{huang2020movienet} that offers a variety of downstream applications including video moderation, search, and recommendation. However, the long-form nature of movies makes labeling of their scenes a laborious process, which limits the effectiveness of traditional end-to-end supervised learning methods for tasks related to automatic scene understanding.

The general problem of learning from limited labels has been explored from multiple perspectives~\cite{wang2020generalizing}, among which contrastive learning~\cite{jaiswal2021survey} has emerged as a particularly promising direction. Specifically, using natural language supervision to guide contrastive learning~\cite{radford2021learning} has shown impressive results specially for zero-shot image-classification tasks. However, these methods rely on image-text pairs which are hard to collect for long-form videos. Another important set of methods within the space of contrastive learning use a \textit{pretext task} to contrast similar data-points with randomly selected ones~\cite{He_2020_CVPR}\cite{chen2020simple}. However, most of the standard data-augmentation schemes~\cite{He_2020_CVPR} used to define the pretext tasks for these approaches have been shown to be not as effective for scene understanding~\cite{Chen_2021_CVPR}. 

\begin{figure*}[!ht]
	%\teaser{
		\centering
		\vspace{-0.3cm}
		\includegraphics[width=0.9\textwidth]{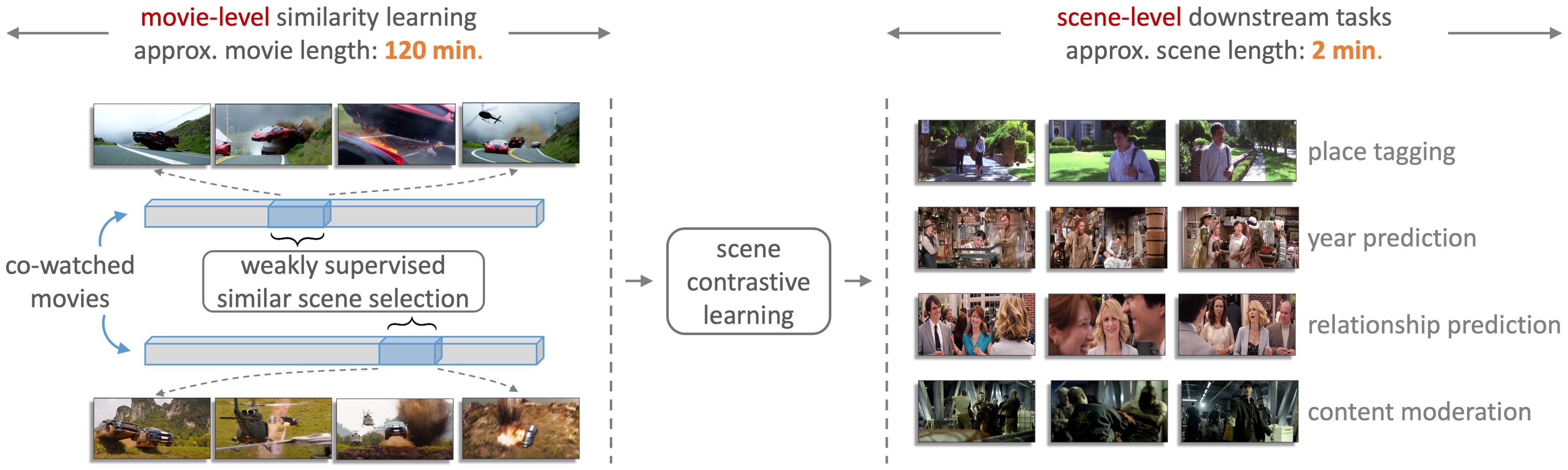}
		\vspace{-0.2cm}\caption{\small {\textbf{Approach Overview -- } We employ commonly available movie metadata (\textit{e.g.}, co-watch, genre, synopsis) to define movie similarities. The figure illustrates a pair of similar movies where movie similarity is defined based on co-watch information, \textit{i.e.}, viewers who watched one movie often watched the second movie as well. Our approach automatically selects thematically similar scenes from such similar movie-pairs and uses them to learn scene-level representations that can be used for a variety of downstream tasks.}}\vspace{-0.2cm}
		\label{fig:teaser_fig}
		%}
\end{figure*}

To address these challenges, we propose a novel contrastive learning approach to find a general-purpose scene representation that is effective for a variety of scene understanding tasks. Our \textbf{key intuition} is that commonly available movie metadata (\textit{e.g.}, co-watch, genre, synopsis) can be used to effectively guide the process of learning a generalizable scene representation. Specifically, we use such movie metadata to define a measure of movie-similarity, and use it during contrastive learning to limit our search for positive scene-pairs to only the movies that are considered similar to each other. This allows us to find positive scene-pairs that are not only visually similar but also semantically relevant, and can therefore provide us with a much richer set of geometric and thematic data-augmentations compared to previously employed augmentation schemes~\cite{He_2020_CVPR}~\cite{Chen_2021_CVPR} (see Figure~\ref{fig:teaser_fig} for illustration). Furthermore, unlike previous contrastive learning approaches that mostly focus on images~\cite{He_2020_CVPR}~\cite{chen2020simple}~\cite{Dwibedi_2021_ICCV} or shots ($\S$~\ref{method} for definition)~\cite{Chen_2021_CVPR}, our approach builds on the recent developments in vision transformers~\cite{dosovitskiy2020vit} to allow using variable-length multi-shot inputs. This enables our method to seamlessly incorporate the interplay among multiple shots resulting in a more general-purpose scene representation.

Using a newly collected internal dataset $\mathsf{MovieCL30K}$ containing $30,340$ movies to learn our scene representation, we demonstrate the flexibility of our approach to handle both individual shots as well as multi-shot scenes provided as inputs to outperform existing state-of-the-art results on diverse downstream tasks using multiple public benchmark datasets~\cite{Wu_2021_CVPR}\cite{huang2020movienet}\cite{rao2020local}. Furthermore, as an important practical application of long-form video understanding, we apply our scene representation to another newly collected dataset $\mathsf{MCD}$ focused on large-scale video moderation with $44,581$ video clips from $18,330$ movies and TV episodes containing $\mathsf{sex}$, $\mathsf{violence}$, and $\mathsf{drug}$-$\mathsf{use}$ activities. We show that learning our general-purpose scene representation is crucial to recognize such age-appropriate video-content where existing representations learned for short-form action recognition or image classification are significantly less effective.
\vspace{-0.0cm}
\section{Related Work}
\label{related}

\vspace{-0.1cm}\noindent\textbf{a. Long-Form Video Understanding:} Recent work on semantic understanding of long-form movies and TV episodes have used multi-shot scenes as their processing unit. For example in MovieNet~\cite{huang2020movienet}, manually annotated multi-shot scenes were used to train various recently proposed models~\cite{carreira2017quo}~\cite{feichtenhofer2019slowfast} to evaluate their performance on multiple tasks related to scene tagging, \textit{e.g.}, recognizing places or actions in those scenes. In~\cite{Liu_2021_CVPR}, multi-shot clips of movies and TV episodes were categorized into $25$ event classes for their temporal localization. The results in~\cite{Liu_2021_CVPR} show that state-of-the-art event localization models~\cite{xu2020g}~\cite{zeng2019graph} do not perform as well on long-form movies and TV episodes compared to their performance on short-form video datasets like THUMOS14~\cite{THUMOS14}. A long-form video understanding (LVU) dataset was recently proposed in~\cite{Wu_2021_CVPR} with nine different tasks related to semantic understanding of video-clips that were cut-out from full-length movies. This work also proposed an object-centric transformer-based video recognition architecture that outperformed SlowFast~\cite{feichtenhofer2019slowfast} and VideoBert~\cite{sun2019videobert} models on their LVU~\cite{Wu_2021_CVPR} dataset. To complement existing datasets focusing on regular everyday activity-categories, we 
%propose
collect a new dataset focusing on video moderation of sensitive activities including $\mathsf{sex}$,  $\mathsf{violence}$, and $\mathsf{drug}$-$\mathsf{use}$, and show how our scene representation can be applied to recognize these activities.

\vspace{0.075cm} \noindent\textbf{b. Contrastive Learning:} As an important subset of self-supervised learning~\cite{jing2020self}, contrastive learning~\cite{le2020contrastive} attempts to learn data representations by contrasting similar data against dissimilar data while using a contrastive loss. Recent approaches for contrastive learning~\cite{He_2020_CVPR}~\cite{chen2020simple} have successfully been extended for a variety of applications in image classification~\cite{tian2020makes}~\cite{henaff2020data}~\cite{grill2020bootstrap} as well as video understanding~\cite{han2020memory}~\cite{Chen_2021_CVPR}~\cite{Zolfaghari_2021_ICCV}. More recently, various visio-linguistic models~\cite{radford2021learning} have adopted natural language supervision during contrastive learning to learn correspondences between image and text pairs, which has achieved impressive results especially for zero-shot image classification tasks. Most of these approaches however focus on images or short-videos, and our experiments show that they do not perform as well on tasks related to long-form video understanding.

\vspace{0.075cm} \noindent\textbf{c. Vision Transformer:} Derived from the work on self-attention~\cite{vaswani2017attention}, transformers have been extensively studied in NLP~\cite{wolf2020transformers}~\cite{devlin2018bert}. Building on this line of work, recently proposed vision transformer (ViT)~\cite{dosovitskiy2020vit} has successfully exceeded state-of-the-art results when pre-trained on large-scale datasets. The data-efficiency of ViT was recently improved in~\cite{touvron2021training} by using token-based distillation for model training.
More recently, ViT-based architectures were adopted for video recognition, \textit{e.g.}, TimeSformer~\cite{bertasius2021space}, ViViT~\cite{arnab2021vivit}, and MViT~\cite{Fan_2021_ICCV}.
These models take images or short-form videos as inputs, and therefore cannot be effectively applied to longer videos without additional tuning.

\vspace{-0.0cm}
\begin{figure*}[!ht]
	\centering
	\includegraphics[width=1.0\textwidth]{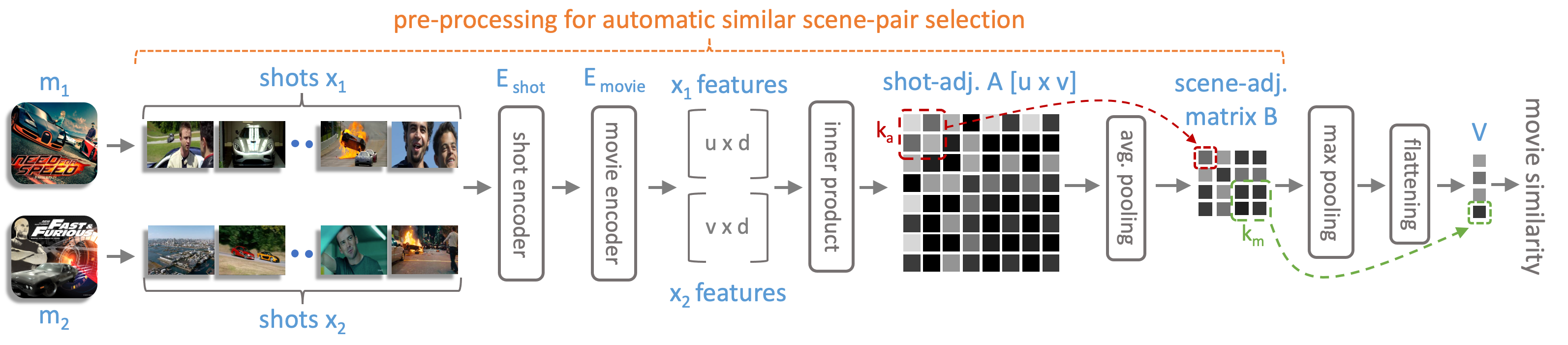}
	\vspace{-0.7cm}
	\caption{\small{\textbf{Movie Similarity Learning -- } Shots of a movie-pair $\mathbf{m}_1$ and $\mathbf{m}_2$ are first provided to $\mathbf{E}_{\textrm{shot}}$ and $\mathbf{E}_{\textrm{movie}}$ to extract $\mathbf{d}$-dimensional features with numbers of $\mathbf{u}$ and $\mathbf{v}$ shots, respectively. We take the dot-product of shot-feature matrices followed by successive pooling and flattening operations to get an output vector $\textbf{V}$, which is then regressed against defined movie-level similarity $\mathcal{S}$ between $\mathbf{m}_1$ and $\mathbf{m}_2$.}}\vspace{-0.0cm}
	\label{pipeline_1}
\end{figure*}

\begin{figure}[!ht]
	\centering
	\includegraphics[width=0.48\textwidth]{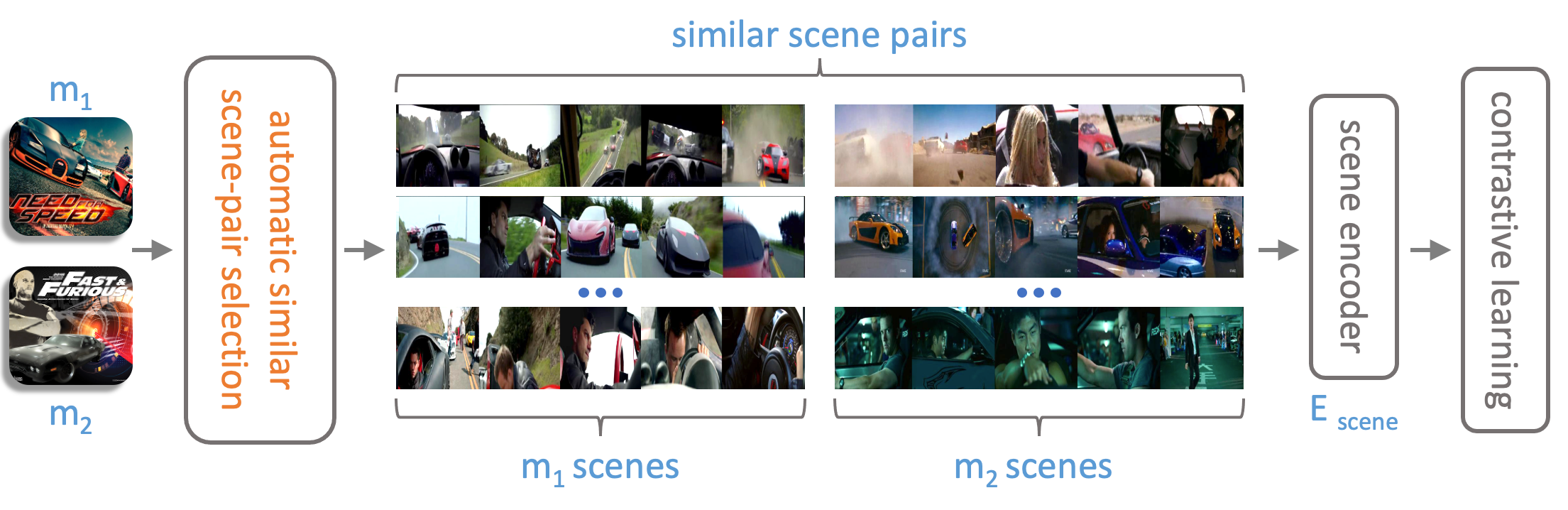}
	\vspace{-0.7cm}
	\caption{\small{\textbf{Contrastive Scene Representation Learning -- } Given a pair of similar movies $\mathbf{m}_1$ and $\mathbf{m}_2$ as determined by $\mathcal{S}$, we use a set of pre-processing operations (as shown in Figure~\ref{pipeline_1}) to select similar scene-pairs (see supplementary material for more examples). This process is applied to all similar movie-pairs to get a set of paired scenes that is used for scene-level contrastive learning.}}\vspace{-0.5cm}
	\label{pipeline_2}
\end{figure}

\vspace{-0.0cm}
\section{Method}
\label{method}

\noindent For consistency, let us define a \textbf{shot} as a series of frames captured from the same camera over a consecutive period of time~\cite{sklar_film}. We define a \textbf{scene} as a series of consecutive shots without requiring manually annotated scene-boundaries. Note that our definition of scenes is less constrained than what has previously been used~\cite{huang2020movienet}~\cite{rao2020local}\cite{Chen_2021_CVPR}, and enables our approach to work seamlessly with settings that do~\cite{huang2020movienet} or do not have~\cite{Wu_2021_CVPR} scene boundaries available.

Our approach consists of four key steps. First, we train a shot encoder to provide appearance-based representations using unlabeled movie-shots. Second, we use commonly available movie metadata (\textit{e.g.}, co-watch, genre or synopsis) to define a movie-level similarity $\mathcal{S}$ (see $\S$~\ref{mcl30k} for details). We use $\mathcal{S}$ as pseudo-labels to train a movie-level encoder that maximizes the similarity between scenes from similar movies (as defined by $\mathcal{S}$). Third, we use this trained movie-level encoder to select similar scene-pairs for contrastive learning of scene encoder. Lastly, we use the learned scene representation for downstream tasks.

\vspace{-0.0cm}
\subsection{Movie Similarity Learning}

\vspace{-0.0cm}
\subsubsection{Shot Encoder}
\label{method_shot}

\noindent As it is generally inefficient to directly train a network with all shots from movies as input in an end-to-end manner~\cite{rao2020local}\cite{huang2020movienet}\cite{Chen_2021_CVPR}, we divide movie-level training into two stages, \textit{i.e.},: (a) shot encoder training, and (b) movie encoder training. During the first stage, we train a shot encoder end-to-end to provide features mostly focusing on appearance-related information without much contextual meaning.

%\noindent Given the long-form nature of full-length movies, it is generally not efficient to directly train a network with all shots from movies as input in an end-to-end manner~\cite{rao2020local}\cite{huang2020movienet}\cite{Chen_2021_CVPR}. To this end, we divide the movie-level training into two stages: shot encoder training and movie-level encoder training, respectively. During the first stage, we train a shot encoder end-to-end that can provide features mostly focusing on appearance-related information without much contextual semantic meaning.
%Then, the shot encoder is fixed during movie-encoder learning for better efficiency.

Given a query shot $\mathbf{x_{t}}$ at time $t$, we use its neighboring shots $\mathbf{x_{t-2}}$ and $\mathbf{x_{t+2}}$ as a pair of potential positive keys. We train our shot encoder to distinguish a positive key shot from randomly selected negative key shots, where the positive logits are defined as:
\begin{equation}
\small{
\label{eq_shot}
\mathbf{\max(E_{shot}(x_{\smaller{t}}) \cdot E_{shot}(x_{\smaller{t\textrm{-}2}}), E_{shot}(x_{\smaller{t}}) \cdot E_{shot}(x_{\smaller{t\textrm{+}2}}))}
}
\end{equation}

\vspace{-0.5cm}
\subsubsection{Movie Encoder}
\label{method_movie}

\noindent After training the shot encoder, we fix it to extract shot representations and add a light-weight encoder on top it as movie encoder. As illustrated in Figure~\ref{pipeline_1}, we divide a given movie-pair $\mathbf{m}_1$ and $\mathbf{m}_2$ into their constituent shots~\cite{sidiropoulos2011temporal} $\mathbf{x}_1$ and $\mathbf{x}_2$ respectively. We extract features of these shots using the fixed shot encoder $\mathbf{E}_{\textrm{shot}}$, so that $\mathbf{x}_1$ and $\mathbf{x}_2$ are represented by two feature matrices $\mathbf{E}_{\textrm{shot}}(\mathbf{x}_1)$ and $\mathbf{E}_{\textrm{shot}}(\mathbf{x}_2)$ respectively, which are passed through an encoder with learnable parameters $\mathbf{E}_{\textrm{movie}}$, followed by their dot-product to create a shot-adjacency matrix $\mathbf{A}_{\mathbf{x}_1, \mathbf{x}_2}$, \textit{i.e.},
\begin{equation}
\label{eq_movie}
\mathbf{A}_{\mathbf{x}_1, \mathbf{x}_2} = \mathbf{E}_{\textrm{movie}}(\mathbf{E}_{\textrm{shot}}(\mathbf{x}_1)) \bigcdot \mathbf{E}_{\textrm{movie}}(\mathbf{E}_{\textrm{shot}}(\mathbf{x}_2))
\end{equation}

\noindent Note that entries in $\mathbf{A}_{\mathbf{x}_1, \mathbf{x}_2}$ represent pair-wise similarities between shots in movies $\mathbf{m}_1$ and $\mathbf{m}_2$.

Using $\mathbf{k}_\textrm{a}$ contiguous shots as scenes, we convert the shot-adjacency matrix $\mathbf{A}_{\mathbf{x}_1, \mathbf{x}_2}$ into a scene-adjacency matrix $\mathbf{B}_{\mathbf{x}_1, \mathbf{x}_2}$ by applying average pooling with kernel size $\mathbf{k}_\textrm{a}$ and stride $\mathbf{s}_\textrm{a}$ on $\mathbf{A}_{\mathbf{x}_1 \mathbf{x}_2}$ to calculate the average values in each $\mathbf{k}_\textrm{a} \times \mathbf{k}_\textrm{a}$ window in $\mathbf{A}_{\mathbf{x}_1, \mathbf{x}_2}$. Furthermore, we apply max pooling with kernel size $\mathbf{k}_\textrm{m}$ and stride $\mathbf{s}_\textrm{m}$ on $\mathbf{B}_{\mathbf{x}_1, \mathbf{x}_2}$ followed by flattening the output to get the vector $\mathbf{V}_{\mathbf{x}_1, \mathbf{x}_2}$. Intuitively, this step finds the most similar scene-pairs between $\mathbf{x}_1$ and $\mathbf{x}_2$, where each value in $\mathbf{V}_{\mathbf{x}_1, \mathbf{x}_2}$ represents the highest similarity of the scene-pair in a neighborhood of $\mathbf{k}_\textrm{m} \times \mathbf{k}_\textrm{m}$ scenes in $\mathbf{B}_{\mathbf{x}_1, \mathbf{x}_2}$. Finally, we learn the projection between $\mathbf{V}_{\mathbf{x}_1, \mathbf{x}_2}$ and the output using layer $\mathbf{L}_{\textrm{output}}$, where our goal is to predict if $\mathbf{x}_1$ and $\mathbf{x}_2$ are similar at movie-level. That is, if $\mathbf{x}_1$ and $\mathbf{x}_2$ are shots from similar movies as defined by $\mathcal{S}$, our target label is $1$, and $0$ otherwise. During training, we use cross-entropy loss to update $\mathbf{E}_{\textrm{movie}}$ and $\mathbf{L}_{\textrm{output}}$ while $\mathbf{E}_{\textrm{shot}}$ remains unchanged.

A useful way to interpret our movie similarity learning step is to view it from a multiple instance learning (MIL) perspective~\cite{carbonneau2018multiple}. The shot-adjacency matrix $\mathbf{A}_{\mathbf{x}_1, \mathbf{x}_2}$ can be considered as a \textit{bag}, while similar scene-pairs can be thought of as positive instances. Unlike some recent MIL-based approaches~\cite{Angles_2021_CVPR}~\cite{Feng_2021_CVPR}~\cite{Li_2021_CVPR}, we simplify our learning process by adopting standard network operations and loss function instead of specially designed ones, which allows us to use existing well-engineered implementations to get better efficiency and scalability in practice.

\vspace{-0.0cm}
\subsection{Scene Contrastive Learning}
\label{method_scene}

\noindent Given a similar movie-pair as defined by $\mathcal{S}$, we apply $\mathbf{E}_{\textrm{movie}}(\mathbf{E}_{\textrm{shot}}(\cdot))$ to their constituent shots, followed by successive pooling operations to find the scene adjacency matrix $\mathbf{B}$, which is used to
%select the top $50\%$
rank the similarity of scene-pairs. We apply this selection process to all pairs of similar movies to get a collection of paired scenes $\mathbf{P}_\textrm{scene}$ that is used for scene-level contrastive learning. Our contrastive learning approach is shown in Figure~\ref{pipeline_2}, and explained below.

\vspace{-0.0cm}
\subsubsection{Scene Encoder}
\label{method_scene_encoder}

As the inputs to our encoder $\mathbf{E}_{\textrm{scene}}$ for scene contrastive learning are multi-shot sequences, it is important to design $\mathbf{E}_{\textrm{scene}}$ so that it can effectively model the various relationships among input shots. To this end, we build on recent work of ViT~\cite{dosovitskiy2020vit}, and propose a transformer based $\mathbf{E}_{\textrm{scene}}$ that treats patches in input shots as tokens.

Specifically, following~\cite{dosovitskiy2020vit}~\cite{radford2021learning}, for a $k$-frame shot of dimension $(k, w, h, c)$, we first divide it into a sequence of $(k, \frac{w}{p}, \frac{h}{p}, \mathit{c})$ patches, where $(p, p)$ is the size of each patch. To input the shot to a transformer with latent vector size of $\textrm{D}$ dimensions, we apply $\textrm{D}$$\times$$\mathit{c}$ convolutional kernels with kernel size $(\mathit{p}, \mathit{p})$ and stride $(\mathit{p}, \mathit{p})$ to the $(k, \frac{w}{p}, \frac{h}{p}, c)$ patches. This converts the shot into patch embeddings with dimension $(k, \frac{h}{p}, \frac{w}{p}, \textrm{D})$, which is further flattened to a $(k, \textrm{N}, \textrm{D})$-dimensional tensor where $\textrm{N}=(w \cdot h) / p^2$. Furthermore, we prepend a learnable embedding to the patch embeddings similar to the class token used in~\cite{devlin2018bert}. After permutation, the result is $(\textrm{N}$$+$$1, \textrm{D})$-dimensional patch embeddings for each of the $k$ frames. We add $(\textrm{N}$$+$$1, \textrm{D})$-dimensional positional embeddings to patch embeddings to retain positional information, and pass them to successive multi-headed self-attention (MSA) layers as done in~\cite{dosovitskiy2020vit}.

Similarly, for the input of a scene with $n$ shots, we first divide it into a sequence of $(n \cdot k, \frac{\mathit{w}}{\mathit{p}}, \frac{\mathit{h}}{\mathit{p}}, c)$ patches. After convolution and flattening, we get $(\textrm{N} \cdot n \cdot k, \textrm{D})$-dimensional patch embeddings. Notice that this is different from the dimension of a frame, and does not match with the dimension of positional embeddings. Inspired by~\cite{dosovitskiy2020vit} where they mentioned that positional embeddings can be interpolated to take input of higher resolutions, we propose to generalize this property from frame to shot and scene. That is, we interpolate the $\textrm{N}$-dimensional positional embeddings to $\textrm{N} \cdot n \cdot k$, excluding the $1$ dimension corresponding to class token, and add the interpolated positional embeddings to patch embeddings before providing them to MSA layers.

%\vspace{0.1cm} \noindent This operation offers two \textbf{key advantages}:

%\vspace{0.1cm}\noindent \textbf{a. Use of Pre-trained Models:} Note that ViT~\cite{dosovitskiy2020vit} offers impressive performance when trained on large-scale datasets ($14$M-$300$M images). Available datasets for long-form video understanding are of significantly smaller scale compared to large-scale image datasets. Our proposed way to apply $2$-D interpolation to position embeddings allows us to scale them from single frame to shots and scenes, and enables us to readily adopt powerful models~\cite{radford2021learning} pre-trained on large image datasets for long-form video understanding without explicit temporal modeling as required in~\cite{bertasius2021space}~\cite{arnab2021vivit}.

%\vspace{0.1cm}\noindent \textbf{b. Flexibility for Downstream Tasks:}
%\vspace{0.1cm}\noindent \textbf{b. Flexibility and Efficiency:} Using $2$-D interpolation to the position embeddings of our encoder allows it to take variable-length shot-sequences as inputs. This enables our approach to be applicable to the common setting where the lengths of multi-shot sequences in contrastive learning are different from those available in downstream tasks. Moreover, as using $2$-D interpolation to the position embeddings of our encoder does not add extra computation along the temporal dimension, it allows us to perform training efficiently which is particularly important for large-scale settings.

\vspace{0.0cm} This operation offers flexibility and efficiency for downstream tasks. That is, using $2$-D interpolation to the positional embeddings of our encoder allows it to take variable-length shot-sequences as inputs. This enables our approach to be applicable to the common setting where the lengths of multi-shot sequences in contrastive learning are different from those available in downstream tasks. Moreover, as using $2$-D interpolation to the positional embeddings of our encoder does not add extra computation along the temporal dimension, it allows us to perform training efficiently which is particularly important for large-scale settings.

\vspace{-0.2cm}
\subsubsection{Contrastive Learning}
Our scene contrastive learning step follows some of the recent works~\cite{He_2020_CVPR}~\cite{Chen_2021_CVPR} with two key differences: (a) unlike the image or shot-augmentation focused pretext tasks previously used, we define scene-level pretext task that uses commonly available movie metadata making it more effective for long-form video understanding, and (b) our use of ViT-based~\cite{dosovitskiy2020vit} scene encoder allows variable-length inputs and the possibility to adopt large-scale pre-trained models, which is not something that previously used ResNet-based encoders~\cite{Chen_2021_CVPR} could offer.

Specifically, we define our pretext task as a dictionary look-up for the scenes selected at the end of our movie similarity learning step. That is, given a query scene $q$, its positive key scene $k_0$ is determined in $\mathbf{P}_\textrm{scene}$, and the objective is to find $k_0$ among a set of random scenes $\{k_1, k_2, ..., k_K\}$. The problem is converted to a ($K+1$)-way classification task by calculating similarity with dot-product, and we use InfoNCE as the contrastive loss function \cite{oord2018representation}:
\begin{equation}
\label{loss}
\mathcal{L}_{q} = -\textrm{log} \frac{\textrm{exp}(f(q | \theta_q) \cdot g(k_0|\theta_k)/ \tau)}{\sum\limits_{i=0}^{\textrm{K}}\textrm{exp}(f(q | \theta_q) \cdot g(k_i|\theta_k) / \tau)}
\end{equation}
where $f(\cdot | \theta_q)$ is the query encoder with parameters $\theta_q$ updated during back-propagation, $g(\cdot | \theta_k)$ is the key encoder with parameters $\theta_k$ learned by momentum update as done in \cite{He_2020_CVPR}, and $\tau$ is the temperature as illustrated in \cite{wu2018unsupervised}.

During contrastive learning, the scene encoder $\mathbf{E}_{\textrm{scene}}$ is the query encoder $q$ without the last fully-connected layer, and is updated based on the selected similar scenes $\mathbf{P}_\textrm{scene}$. After training converges, $\mathbf{E}_{\textrm{scene}}$ is the outcome from this stage, which can then be used for downstream tasks.

\iffalse
\vspace{-0.0cm}
\subsection{Application to Downstream Tasks}
\label{method_down}

\noindent We train encoder $\mathbf{E}_{\textrm{scene}}$ as detailed above and apply it to downstream tasks as a general-purpose scene-level encoder. Specifically, depending on the downstream task, we first process the video-data to a sequence of shots or frames. Then, we extract features for each sample by $\mathbf{E}_{\textrm{scene}}$, which then can be used as input for downstream classifiers.
\fi
\vspace{-0.0cm}
\section{Experiments}
\label{exp}

\noindent We first discuss one of our newly collected movie datasets $\mathsf{MovieCL30K}$ that we used to learn our scene representation followed by describing the implementation details of our algorithm. We then present comparative results on multiple benchmark datasets~\cite{Wu_2021_CVPR}~\cite{huang2020movienet}~\cite{rao2020local} demonstrating significant gains of our approach over existing state-of-the-art models for a diverse set of downstream tasks in LVU~\cite{Wu_2021_CVPR} and MovieNet~\cite{huang2020movienet}. To further demonstrate the generalizability of our learned scene representation, we test it on another newly collected dataset $\mathsf{MCD}$ focused on large-scale moderation of age-sensitive activities
%including $\mathsf{sex}$,  $\mathsf{violence}$, and $\mathsf{drug}$-$\mathsf{use}$
and show substantial gains over existing state-of-the-art approaches.

\iffalse
\setlength{\tabcolsep}{2pt}
\begin{table}[t]
	\begin{center}
		\smallest{
			\begin{tabu}{c|ccccccccccc}
				\hline
				genre	&action	&drama	&comedy	&horror	&doc	&foreign	&animation	&kids	&sci-fi	&romance	&other \\\hline
				\%	&27.4	&21.6	&17.5	&6.5	&4.6	&4.6	&4.1	&3.1	&2.8	&1.2	&6.1 \\\hline
			\end{tabu}
		}
	\end{center}
	\vspace{-0.2cm}\caption{\small{Distribution of movie genres in $\mathsf{MovieCL30K}$ dataset.}}\vspace{-0.0cm}
	\label{dist_genre}
\end{table}
\setlength{\tabcolsep}{4pt}\textbf{}
\fi

\vspace{-0.0cm}
\subsection{Dataset and Implementation}
\subsubsection{MovieCL30K}
\label{mcl30k}
%\noindent For contrastive learning of our scene-representation, w
\noindent We compiled a new dataset called $\mathsf{MovieCL30K}$ with $30,340$ movies from $11$ genres with the genre-distribution of [action, drama, comedy, horror, doc, foreign, animation, kids, sci-fi, romance, other] being [27.4\%, 21.6\%, 17.5\%, 6.5\%, 4.6\%, 4.6\%, 4.1\%, 3.1\%, 2.8\%, 1.2\%, 6.1\%].
%(see Table~\ref{dist_genre} for genre-distribution).
%Similar to some other internal datasets, e.g., JFT-300M~\cite{sun2017revisiting} and CLIP-400M~\cite{radford2021learning}, we will not be able to publicly release $\mathsf{MovieCL30K}$ and $\mathsf{MCD}$ due to copyright and other constraints, but we will explore other possibilities to further contribute to the community by, e.g., releasing pre-trained model weights.
%Thus, we do not claim dataset collection as a contribution in this paper.
%To compute movie-similarity in this data, we used
%readily accessible movie information available on IMDb
We empirically compare commonly available movie metadata including: (i) co-watch, (ii) genre, and (iii) synopsis.
%More-like-this is a pre-computed comprehensive similarity measure between movies on IMDb which takes into account multiple factors including genres, country-of-origin and actors.

We collect movie co-watch information based on the watch history of movies. That is, for each movie $\mathbf{m}_1$, we keep records of movies that are watched after $\mathbf{m}_1$, and then rank the records to build a set of movies that are most frequently watched after $\mathbf{m}_1$. %For each movie in $\mathsf{MovieCL30K}$, we select its most co-watched movies based on their rankings.
To use genre to find movies similar to an input movie, we randomly select movies with the same genre as the input movie. Lastly, to use synopsis for movie-level similarity, we first extract textual-embeddings of synopsis using a pre-trained natural language processing model~\cite{liu2019roberta}, and then compute pair-wise movie-similarities as the dot-product of their textual-embeddings followed by selecting the closest movies. When using any of the aforementioned movie-similarity measures, we select three most similar movies on average for each movie in $\mathsf{MovieCL30K}$.

\subsubsection{Implementation Details}
\noindent We divide movies in $\mathsf{MovieCL30K}$ into their constituent $\sim$$33$ million shots, and only keep one frame per shot for efficiency. We first use ViT~\cite{dosovitskiy2020vit} as our shot encoder $\mathbf{E}_{\textrm{shot}}$ and train it with $30\%$ of all shots forming query and key pairs. Then we fix and use $\mathbf{E}_{\textrm{shot}}$ to extract representations of all shots as shown in Figure \ref{pipeline_1}. The movie-level encoder $\mathbf{E}_{\textrm{movie}}$ comprises of two fully-connected layers each with $512$ dimensions, and a dropout layer with $0.5$ probability between them.
%The fact that each movie in $\mathsf{MovieCL30K}$ consists more than $1,000$ shots on average leads us to keep our $\mathbf{E}_{\textrm{shot}}$ constant and to only update parameters of $\mathbf{E}_{\textrm{movie}}$ during movie-level training to help us achieve high efficiency.
We keep length of each movie to be $1,024$ shots, and zero-pad the movies shorter than that. Kernel size for both max and average pooling are set to $16$, with a stride of $8$.

After movie-level similarity learning (Figure~\ref{pipeline_1}), we only keep the top $50\%$ of most similar scene-pairs selected in each movie-pair because we empirically found that the scene-pairs can get noisier and become less informative beyond this ratio.
%and use a scene-length of $9$ shots for contrastive learning (Figure~\ref{pipeline_2}).
We follow the implementation in~\cite{He_2020_CVPR} for momentum contrastive learning and use ViT~\cite{dosovitskiy2020vit} as our scene encoder instead of ResNet~\cite{he2016deep} used in~\cite{He_2020_CVPR}. Moreover, we optimize with AdamW \cite{loshchilov2019decoupled} instead of SGD~\cite{goyal2017accurate}.
%Unlike shot-level learning where we directly use ViT~\cite{dosovitskiy2020vit}, 
For scene representation learning, we interpolate the positional embeddings (see $\S$~\ref{method_scene_encoder}) so that the encoder can take variable-length multi-shot sequences as input.

For supervised learning on downstream tasks, unless otherwise specified, we use a simple multilayer perceptron (MLP) with two hidden layers, where the output layer is modified for each various classification or regression task.

\setlength{\tabcolsep}{2pt}
\begin{table*}[t]
	\begin{center}
	\smaller{
		\begin{tabu}{ccccccccc|[1.25pt]ccccccc|cc}
		%\taburowcolors[1]10{lightb..white}
			\hline
			&pre-train   &shot &scene  &\# of shots   &\multirow{2}{*}{backbone} &hyper-params   &unit   &\multirow{2}{*}{metadata}    &\multicolumn{7}{c|}{classification $\uparrow$}	&\multicolumn{2}{c}{regression $\downarrow$}\\
			&movies      &enc.   &init.  &-frames  &  &bat.-res. &time    &   &place	&director	&relation	&speak	&writer	&year	&genre	&view	&like\\\tabucline[1.25pt]{-}
			\multicolumn{18}{c}{Various pre-training scales}\\\hline
			1	&30K        &random &random &9-1    &ViT-B/16 &128-128   &1 &co-watch   &63.2	&69.8	&68.7	&40.1	&60.1	&54.1	&55.5	&2.62	&0.161\\
			2	&15K        &random &random &9-1    &ViT-B/16 &128-128   &0.5 &co-watch   &59.1	&66.3	&63.3	&37.5	&55.9	&50.3	&52.7	&3.02	&0.254   \\
			3	&7.5K       &random &random &9-1    &ViT-B/16 &128-128   &0.25 &co-watch   &55.7	&60.9	&59.4	&35.8	&52.7	&48.3	&48.8	&3.19	&0.307   \\
			4	&3K        &random &random &9-1    &ViT-B/16 &128-128   &0.1 &co-watch   &45.3	&49.5	&52.2	&34.6	&47.6	&41.2	&43.6	&3.47	&0.352\\
			5	&300        &random &random &9-1    &ViT-B/16 &128-128   &0.01 &co-watch   &31.2	&29.8	&35.7	&27.5	&30.1   &28.9	&34.3	&4.55	&0.496\\\hline
			\multicolumn{18}{c}{Different numbers of shots and frames}\\\hline
			6	&30K        &random &random &1-1    &ViT-B/16 &128-128   &1 &co-watch   &46.5	&48.2	&44.1	&30.3	&31.1	&34.4	&45.8	&3.89	&0.401\\
			7	&30K        &random &random &4-1    &ViT-B/16 &128-128   &1 &co-watch   &55.6	&60.1	&57.2	&33.4	&50.3	&49.8	&51.1	&3.07	&0.246\\
			8	&30K        &random &random &16-1    &ViT-B/16 &128-128   &1 &co-watch   &62.4	&66.2	&65.3	&39.7	&56.2	&54.3	&52.4	&2.93	&0.173\\
			9	&30K       &random &random &9-3    &ViT-B/16 &128-128   &3 &co-watch   &63.5	&70.1	&69.3	&40.4	&59.7	&54.9	&54.7	&2.59	&0.159\\\hline
			\multicolumn{18}{c}{Different similarity measures}\\\hline
			10	&30K        &random &random &9-1    &ViT-B/16 &128-128   &1 &random   &49.2	&48.8	&51.5	&34.6	&37.2	&36.7	&49.1	&3.94	&0.454\\
			11	&30K        &random &random &9-1    &ViT-B/16 &128-128   &1 &genre   &62.9	&58.1	&66.3	&39.5	&47.3	&45.2	&56.2	&3.85	&0.351\\
			12	&30K        &random &random &9-1    &ViT-B/16 &128-128   &1 &synopsis   &56.7	&54.1	&65.3	&36.8	&45.9	&34.5	&50.7	&3.87	&0.471\\
			13	&30K        &random &random &9-1    &ViT-B/16 &128-128   &1 &gen.+syn.   &64.1	&65.5	&69.6	&41.5	&52.3	&51.1	&56.7	&2.94	&0.199\\\hline
			\multicolumn{18}{c}{Comparisons on weight initialization and backbone}\\\hline
			14	&30K        &CLIP   &CLIP  &9-1     &ViT-B/16 &128-128   &1 &co-watch   &61.2 &65.3	&67.4	&43.7	&54.1	&53.5	&56.3 &2.75	&0.178\\
			15	&3K         &random &random &9-1    &MViTv2-S &32-224   &1 &co-watch   &47.6	&51.2	&51.3	&35.3	&45.1	&44.7	&43.7	&3.51	&0.343\\\hline
			
		\end{tabu}
		}
	\end{center}
	\vspace{-0.6cm}\caption{\small{Ablation study results on validation-set of benchmark LVU dataset~\cite{Wu_2021_CVPR}. Note that for classification tasks higher is better, and for regression tasks lower is better.}}\vspace{-0.0cm}
	\label{exp_ablation}
\end{table*}
\setlength{\tabcolsep}{4pt}

\setlength{\tabcolsep}{2pt}
\begin{table*}[t]
	\begin{center}
	\smaller{
		\begin{tabu}{cccccc|[1.25pt]ccccccc|cc}
			\hline
			&\multirow{2}{*}{models}   &\multirow{2}{*}{pre-train data} &\multirow{2}{*}{modalities} &frames &params   &\multicolumn{7}{c|}{classification $\uparrow$}	&\multicolumn{2}{c}{regression $\downarrow$}\\
			&   &      &   &/scene  &sup. learn &place	&director	&relation	&speak	&writer	&year	&genre	&view	&like\\\tabucline[1.25pt]{-}
			\multicolumn{15}{c}{End-to-end learning approaches}\\\hline
			1	&Video Bert \cite{sun2019videobert}	&30K LVU movie clips   &visual   &60    &$\sim$8.77M    &54.9	&47.3	&52.8	&37.9	&38.5	&36.1	&51.9	&4.46	&0.320\\
			2	&SlowFast R101 \cite{feichtenhofer2019slowfast}	&30K LVU movie clips   &visual   &60    &$\sim$44M    &54.7	&44.9	&52.4	&35.8	&36.3	&52.5	&53	&3.77	&0.386\\
			3	&OT \cite{Wu_2021_CVPR}	&30K LVU movie clips   &visual   &60    &$\sim$27M  &56.9	&51.2	&53.1	&39.4	&34.5	&39.1	&54.6	&3.55	&0.230\\
			4   &ViS4mer \cite{islam2022long}	&30K LVU movie clips    &visual &60   &$\sim$3.6M   &67.4	&62.6	&57.1	&40.7	&48.8	&44.7	&54.7	&3.63	&0.260\\
			\multirow{2}{*}{5}   &Hierarchical	&240K Kinetics &\multirow{2}{*}{visual} &\multirow{2}{*}{16} &\multirow{2}{*}{$\sim$27M}    &\multirow{2}{*}{44.1}	&\multirow{2}{*}{40.1}	&\multirow{2}{*}{50.9}	&\multirow{2}{*}{34.1}	&\multirow{2}{*}{31.4}	&\multirow{2}{*}{29.6}	&\multirow{2}{*}{51.1}	&\multirow{2}{*}{4.88}	&\multirow{2}{*}{0.353}\\
			&OT \cite{Xiao_2022_CVPR}	&+23.5K VidSitu & & & & & & & & & & & & \\\hline
			\multicolumn{15}{c}{Representation-based approaches}\\\hline
			6	&CLIP \cite{radford2021learning}	&400M image-text pairs   &vis.+text &9  &$\sim$0.7M   &52.9	&56.2	&56.1	&36.7	&37.8	&46.4	&50.9	&3.85	&0.411\\
			7   &Merlot Reserve \cite{zellers2022merlotreserve}	&20M Youtube videos    &vis.+text+aud.   &8   &$\sim$0.7M  &59.2   &54.4   &60.0   &41.1   &38.5 &49.7 &54.6    &3.03  &0.217\\
			8   &ShotCoL \cite{Chen_2021_CVPR}	&2.5M movie shot pairs   &visual &9  &$\sim$1.3M &45.3   &49.5   &46.7   &31.1   &29.8   &36.7   &43.1   &4.79   &0.397\\
			9   &Bridge Former \cite{ge2022bridgeformer}	&3.3M image+2.5M video  &vis.+text &4  &$\sim$0.4M  &62.8   &55.7   &60.4   &40.9   &49.7   &41.4   &52.9   &3.97   &0.312\\
			10	&Ours       &2.5M movie scene-pairs &visual &9  &$\sim$0.7M   &\textbf{68.2}	&\textbf{70.9}	&\textbf{71.2}	&\textbf{42.2}   &\textbf{53.7}	&\textbf{57.8}	&\textbf{55.9}   &\textbf{2.79}   &\textbf{0.192}\\
			&   &   &   &   &   &\color{Green}\smallest{\textbf{+0.8}}	&\color{Green}\smallest{\textbf{+8.3}}	&\smallest{\textbf{\color{Green}+10.8}}	&\smallest{\textbf{\color{Green}+1.1}}   &\smallest{\textbf{\color{Green}\textbf{+4.0}}}	&\smallest{\textbf{\color{Green}+5.3}}	&\smallest{\textbf{\color{Green}+1.2}}   &\smallest{\textbf{\color{Green}-0.24}}   &\smallest{\textbf{\color{Green}-0.025}}\\\hline
		\end{tabu}
		}
	\end{center}
	\vspace{-0.5cm}\caption{\small{Quantitative comparisons on test-set of benchmark LVU dataset~\cite{Wu_2021_CVPR}. Our approach is evaluated on nine diverse tasks and compared against multiple state-of-the-art models. Note that for classification tasks higher is better, and for regression tasks lower is better.}}\vspace{-0.1cm}
	\label{exp_lvu}
\end{table*}
\setlength{\tabcolsep}{4pt}
%average (-3/67.4+8.7/62.6+10.8/60.4+0.8/41.1+10.4/49.7+3.4/52.5+0.2/54.7)/7, (0.34/3.03+0.054/0.217)/2
%Comparisons of our approach using different movie metadata are also provided. 

\vspace{-0.0cm}
\subsection{Comparisons on LVU Benchmark}

\noindent To ensure that there is no overlap between pre-training and downstream evaluation, during the representation learning step, we exclude movies from $\mathsf{MovieCL30K}$ that have the same IMDb IDs as the ones in validation and test sets of compared benchmark datasets.

\vspace{-0.0cm}
%\subsubsection{LVU Benchmark}
We first evaluate our model using the benchmark LVU dataset~\cite{Wu_2021_CVPR} which contains nine diverse tasks including: place (scene), director, relationship, way-of-speaking, writer, year, genre, view-count, and like-ratio. The total number of labeled videos in LVU dataset~\cite{Wu_2021_CVPR} is $\sim$$11$K with the splits of training, validation, and testing sets to be $70$\%, $15$\%, and $15$\%, respectively. Following~\cite{Wu_2021_CVPR}, view-count and like-ratio are evaluated by mean-squared-error in regression (lower is better), while other seven tasks are evaluated using top-$1$ accuracy in classification (higher is better).
%Note that unlike $\mathsf{MovieCL30K}$ dataset where each movie can be up to $120+$ minutes long, LVU dataset contains video clips that are generally of a few minute duration.

\subsubsection{Ablation Study}
To assess the effectiveness of different components in our model, we present the ablative comparisons in Table~\ref{exp_ablation} regarding factors including: 1) number of movies used during pre-training, 2) whether the shot encoder is initialized using CLIP~\cite{radford2021learning} visual encoder and kept fixed or randomly initialized and trained by our method, 3) whether the scene encoder is initialized with CLIP~\cite{radford2021learning} weights or randomly, 4) influence of number of shots in each scene and number of frames in each shot, 5) different backbone architectures, 6) hyper-parameters including batch size and input resolution, 7) how long it takes for the training to complete compared to row 1 as one unit time, 8) the metadata used to define similar movie-pairs. Results on validation-set of nine tasks in LVU~\cite{Wu_2021_CVPR} are presented for each setting. The insights from these comparisons are summarized as follows, where the results are compared against row 1 as the base case.

\vspace{0.1cm}\noindent \textbf{a. Training scale is critical for model performance.} From rows 1 to 5 in Table~\ref{exp_ablation}, we compare different scales of pre-training. We can see that our model can work with as little as $3\textrm{K}$ pre-training movies, but it takes about $15\textrm{K}$ movies for our method to compare favorably with previous state-of-the-art results~\cite{Wu_2021_CVPR}. Since we do not need any manually labeled data for pre-training, we can scale our approach for better performance relatively more easily.
%, and this aligns with the trend reported in other large-scale representation learning approaches~\cite{radford2021learning}\cite{zellers2022merlotreserve}\cite{Chen_2021_CVPR}\cite{ge2022bridgeformer}.
%Meanwhile, we can see that pre-training on movies and TV episodes performs differently on various downstream tasks. It is possibly because co-watched movies can be very filmed independently and thus do not necessarily have same settings while TV episodes can be produced in the same setting and thus provide similar scene pairs that are more visually similar. 

\vspace{0.1cm}\noindent \textbf{b. Number of shots is important.} From rows 6 and 7, comparing with row 1, the performance would be improved when we increase the number of shots per scene. This is reasonable since we can incorporate more information with more shots.
%The training unit time stays roughly the same mainly because with shorter length of scenes, we can find more scene pairs from same amount of movies, but even with more scene pairs, it is not directly contributing to the performance on downstream tasks.
However, when we further increase the number to 16 shots per scene in row 8, the performance drops. This is possibly because with the longer context, some scenes may cross scene boundaries, and thus provide noisy or irrelevant information from different scenes. Also, the number of scene-pairs would decrease with more shots per scene, and thus provide fewer training samples for scene contrastive learning, which may further reduce the effectiveness of the learned scene representations. Adding more frames per shot as in row 9 can also increase the performance. However, when we have more frames to process, it takes significantly more time for training with only marginal performance gain. Thus, we choose to use $9$ shots per scene with $1$ frame per shot for all future experiments as default.

\vspace{0.1cm}\noindent \textbf{c. Movie metadata matters.} In rows 10 to 12, we compare the representations learned using different type of movie metadata. We can see that co-watch, genre and synopsis are all effective measures for movie similarity. Using co-watch can achieve overall better results, which is likely due to the fact that it can incorporate more complex relationships between movies, and thus provide more diverse scene-pairs for representation learning. Moreover, when we concatenate the features separately learned by genre and synopsis, it offers similar or even better accuracy compared to co-watch, which indicates that further improvements can be achieved when combining scene representations learned from different metadata information. However, as shown in row 10, randomly picking movie-pairs from the $30$K movies and applying our approach to them does not offer results that are comparable to what we get when using other types of metadata. This indicates that informative and meaningful metadata is necessary for our model to perform well.
%is able to outperform genre and synopsis when they are  measures. being a more comprehensive similarity measure incorporating multiple factors, is able to outperform other considered similarity measures.

\vspace{0.1cm}\noindent \textbf{d. Our method does not rely on pre-trained weights.} In row 14, we swap our shot encoder to the visual encoder in CLIP~\cite{radford2021learning} and use CLIP weights as initialization for scene encoder. We can see that the performance is comparable to random initialization, showing that our whole framework can be trained from scratch using solely movie metadata.
%The shot encoder we are learning is under the realm of movies, and it is different from ShotCoL~\cite{Chen_2021_CVPR} in two aspects. First, we select positive keys only between two interval shots at timestamps $t-2$ and $t+2$ given query shots at $t$. This=enables us to do online selection while ShotCoL did offline update of positive key selection. This limits the neighborhood size of our approach but enables our method to be more flexible and dynamic. Also, our goal is not learning the representation that can work specifically well for scene boundary detection tasks. Rather, we need a more generic appearance-based shot encoder that is not geared towards contextual relationship within a specified neighborhood.

\vspace{0.1cm}\noindent \textbf{e. Model-complexity and training-scale trade-off.} When comparing with different backbone architecture MViTv2-S~\cite{Fan_2021_ICCV}\cite{li2021improved} in row 15, we notice that while recent architectures can improve performance, it may take significantly more time for training.
%For example, when using the default setting in MViTv2-S,
%which is already a relatively smaller choice in MViTs~\cite{fan2021multiscale},
%it takes much longer time for the training to complete compared to our setting. 
For example, even when the number of parameters in ViT-B/16 ($87.2$M) is more than the one in MViTv2-S ($26.1$M)~\cite{Fan_2021_ICCV}\cite{li2021improved}, the expected time to complete the training of MViTv2-S with their default setting is more than $10\times$ that of ViT-B/16 due to smaller batch size and larger input size. To keep $\sim$$1$ unit time for training of MViTv2-S, we can only use $3\textrm{K}$ pre-training data. It performs better than row 4 using same amount of data but only requires $0.1$ unit time, and it is not comparable with row 1 that can use $10\times$ of data with $1$ unit time. Thus, we keep our setting on row 1 as default.

\subsubsection{Comparing with State-of-the-art Results}
In Table \ref{exp_lvu} we compare our approach with state-of-the-art results on LVU~\cite{Wu_2021_CVPR} test-set. There are two main parts of Table \ref{exp_lvu} where each model is considered as either an end-to-end learning or representation-based approach. For end-to-end methods, they generally use a smaller scale of pre-training data, followed by end-to-end fine-tuning on downstream tasks in LVU. Thus, they require to train a higher number of parameters during supervised training. A higher number of frames per scene is also used in end-to-end models, which enables them to incorporate more information.

For representation-based approaches, we compare with some recent large-scale pre-training models using various scales and modalities of data. We extract features from their fixed visual encoders and learn a simple MLP for downstream tasks, which significantly reduces the number of parameters needed during supervised learning. We can see that our method is significantly different from other existing representation-based approaches in terms of required modalities. That is, most existing methods rely on cross-modality information to learn representations, and this requires a more complex procedure for data process and collection. The only approach using single modality is ShotCoL~\cite{Chen_2021_CVPR}. However, since they learned shot-level representations that are most effective for scene boundary detection tasks, when applied to semantic understanding of content with longer context, they are not performing as well.

Overall, we can see that comparing with both end-to-end and representation-based approaches, our model can comfortably outperform other state-of-the-art results across different tasks, with a large margin on many of them. These comparisons further validate the effectiveness of our learned scene representations.

\subsection{MovieNet Benchmark}

\noindent MovieNet is a large-scale movie understanding dataset with $1,100$ movies covering a variety of tasks~\cite{huang2020movienet}. However, unlike LVU~\cite{Wu_2021_CVPR} where videos are publicly available, the movies in MovieNet are not publicly released yet due to copyright issues. Only keyframes from each shot are provided, which makes tasks that demand high frame-rate (\textit{e.g.}, action recognition) infeasible. We therefore focus our comparisons to two scene understanding related tasks in MovieNet that do not demand particularly high frame-rate, \textit{i.e.}: (i) place tagging, and (ii) scene boundary detection.
%The key difference between these two tasks is that place-tagging focuses more on the holistic understanding of the scene, while scene-boundary-detection requires the understanding of the relationships among multiple-shots. Our results demonstrate that our scene-representation can work well under both these settings and surpasses their existing state-of-the-art approaches.

\vspace{0.1cm}\noindent \textbf{a. Place Tagging:} There are $90$ categories of places with $19.6 \textrm{K}$ place tags in MovieNet~\cite{huang2020movienet}. The place tagging problem is formulated as a multi-label classification task where each manually labeled scene can have multiple place tags. The results are evaluated by mean average precision (mAP) on the test-set as shown in Table~\ref{res_tag}. We compare our approach with two sets of models: (i) fine-tuning models, and (ii) feature-based approaches. The results on fine-tuning are reported by~\cite{huang2020movienet} where they mentioned that standard action recognition models were adopted. For feature-based approaches, we compare our encoder with CLIP visual encoder~\cite{radford2021learning} and ResNet50~\cite{he2016deep} pre-trained on the Place dataset~\cite{zhou2017places}. We can see that our model can significantly outperform all other compared methods on this task.

\setlength{\tabcolsep}{2pt}
\begin{table}[!t]
	\begin{minipage}[!t]{0.46\linewidth}
	\begin{center}
	\vspace{0.1cm}
	\smaller{
		\begin{tabu}{c|c|c}
			\hline
			\multicolumn{2}{c|}{Models}	&mAP	\\\tabucline[1.25pt]{-}
			Fine-	&TSN~\cite{wang2016temporal}	&8.33\\
			tuning&I3D~\cite{carreira2017quo}	&7.66	\\\hline
			&ImgNet~\cite{deng2009imagenet}	&7.04\\
			Feature&Place~\cite{zhou2017places}	&8.76\\
			+MLP&CLIP~\cite{radford2021learning}	&9.26\\
			&Ours	&\textbf{12.17}\\\hline
		\end{tabu}}
	\end{center}
	\vspace{-0.5cm}\caption{\small{Comparisons with state-of-the-art results on MovieNet place scene tagging using methods based on: (a) action recognition models fine-tuned for this task, and (b) commonly used pre-trained representations for generic downstream tasks.}}
	\label{res_tag}
	\end{minipage}
	\quad
	\begin{minipage}[!t]{0.46\linewidth}
	\begin{center}
		\vspace{-0.0cm}
		\smaller{
			\begin{tabu}{c|c|c}
				\hline
				Models &Pre-train data	&AP	\\\tabucline[1.25pt]{-}
				SimCLR~\cite{chen2020simple}	&same-domain	&41.65	\\
				MoCo~\cite{He_2020_CVPR}	&same-domain	&42.51	\\
				ShotCoL~\cite{Chen_2021_CVPR}	&same-domain	&53.37	\\
				SCRL~\cite{Wu_2022_CVPR}	&same-domain	&54.82	\\
				BaSSL~\cite{Mun_2022_ACCV}	&same-domain	&\textbf{57.40}	\\\hline
				LGSS~\cite{rao2020local}	&cross-domain	&47.1	\\
				CLIP~\cite{radford2021learning}	&cross-domain	&47.4	\\
				ShotCoL~\cite{Chen_2021_CVPR}	&cross-domain	&48.4	\\
				Ours	&cross-domain	&\textbf{55.03}	\\\hline
			\end{tabu}
		}
	\end{center}
	\vspace{-0.5cm}\caption{\small{Comparisons with state-of-the-art results on SBD. Our approach outperforms all methods trained in cross-domain and provides competitive results to models trained in same-domain.}}%outperforms most state-of-the-arts even when pre-trained in cross-domain setting.
	\label{exp_sbd}
	\end{minipage}
\vspace{-0.3cm}
\end{table}
\setlength{\tabcolsep}{4pt}

\vspace{0.1cm}\noindent \textbf{b. Scene Boundary Detection:} Scene boundary detection (SBD) is a challenging and important problem for semantic movie understanding~\cite{rao2020local}\cite{Chen_2021_CVPR}.
%Unlike previous approaches that proposed dedicated SBD models~\cite{rao2020local}\cite{Chen_2021_CVPR}\cite{Wu_2022_CVPR}\cite{Mun_2022_ACCV},
We show that our learned general-purpose scene encoder can be seamlessly applied to this task and provide competitive results to state-of-the-art models. Following~\cite{Chen_2021_CVPR}, we use the subset of $318$ movies from MovieNet~\cite{rao2020local}\cite{huang2020movienet} where manually annotated scene boundaries are provided. The training, testing and validation splits are also provided with $190$, $64$ and $64$ movies respectively. The problem is formulated as binary classification to predict if a shot boundary is also a scene boundary, and the results are evaluated by average precision (AP)~\cite{huang2020movienet}\cite{Chen_2021_CVPR}. Following ShotCoL~\cite{Chen_2021_CVPR}, we use a four-shot sequence as a sample where representations for each shot are independently extracted and concatenated.
Notice that in addition to Place Tagging and LVU tasks where our encoder can take multi-shot clips as input, for SBD, our encoder can take individual shots as input which demonstrates the flexibility of our approach for different types of inputs.

We report the AP results of the different compared approaches in Table~\ref{exp_sbd},
%Notice that all compared methods include two components, \textit{i.e.}, an encoder and a predictor. However, the encoder-settings
where approaches are categorized depending on whether the pre-training data comes from the same domain as the downstream task. For example, there are four encoders in LGSS~\cite{rao2020local}
%with features related to visual, audio, action and actor, respectively. These encoders 
pre-trained in supervised manner on datasets that are different from MovieNet~\cite{huang2020movienet}, and thus it is the cross-domain setting. For SCRL~\cite{Wu_2022_CVPR}, BaSSL~\cite{Mun_2022_ACCV} and MoCo~\cite{He_2020_CVPR} implemented in~\cite{Chen_2021_CVPR}, the representations were learned on the full MovieNet~\cite{huang2020movienet} dataset with $1,100$ movies,
%by the pretext task of image augmentation,
and then applied to the subset of $318$ movies on SBD, making it a same-domain setting. ShotCol~\cite{Chen_2021_CVPR} provided results for both settings with a significant performance gap between them.
%This indicates that even with same pretext task, performance on downstream tasks can vary significantly if domain discrepancy exists.

%Notice that our scene encoder can conveniently take a single shot as input and provide effective representation, which outperforms most state-of-the-art results even when used in cross-domain setting. This further validates the flexibility and generalizability of our approach.
%Notice that our scene encoder can conveniently take a single shot as input and provide effective representation. %while performing comparably with other state-of-the-art models that are dedicated for SBD.
%As shown in Table~\ref{exp_sbd}, our approach can outperform other state-of-the-art models pre-trained in cross-domain by a large margin. It can also provide competitive results to models~\cite{Chen_2021_CVPR}\cite{Wu_2022_CVPR}\cite{Mun_2022_ACCV} that are dedicatedly designed for SBD and pre-trained in same-domain.
As shown in Table~\ref{exp_sbd}, our approach outperforms state-of-the-art models pre-trained in cross-domain, and provides competitive results to models~\cite{Chen_2021_CVPR}\cite{Wu_2022_CVPR}\cite{Mun_2022_ACCV} that are dedicatedly designed for SBD and pre-trained in same-domain.

\subsection{Video Moderation}

\iffalse
\begin{figure}[t]
	\centering
	\includegraphics[width=0.5\textwidth]{cd_eg_blur_1}\vspace{-0.2cm}
	\caption{\small{Examples of $3$ types of age-appropriate activities in our data. Sensitive parts of images have been intentionally redacted here. See supplementary materials for more examples.}}\vspace{-0.0cm}
	\label{eg_cd}
\end{figure}
\fi

%\noindent To further demonstrate the generalizability of our scene-representation, we focus on video-moderation particularly to detect $\mathsf{sex}$, $\mathsf{violence}$, and $\mathsf{drug}$-$\mathsf{use}$ activities in movies and TV shows.
\noindent Large-scale video moderation is one of the most pressing challenges faced by video streaming services. However, most existing action recognition datasets (\textit{e.g.}, Kinetics~\cite{kay2017kinetics,carreira2018short}, THUMOS14~\cite{THUMOS14}) do not cover movies and TV episodes well, while existing movie datasets (AVA~\cite{gu2018ava}, MovieNet~\cite{huang2020movienet}) are not including age-sensitive activities. To this end, we focus on video moderation particularly to detect $\mathsf{sex}$, $\mathsf{violence}$, and $\mathsf{drug}$-$\mathsf{use}$ activities  (see supplementary materials for examples) in movies and TV episodes.
%We first go over a newly collected data for large-scale video-moderation in studio-produced content, followed by presenting comparative results using this data.

\setlength{\tabcolsep}{2pt}
\begin{table}[t]
	\begin{center}
		\smaller{
			\begin{tabu}{c|c|ccc|c}
				\hline
				Models & Pre-training data & $\mathsf{sex}$  &	$\mathsf{violence}$ & $\mathsf{drug}$-$\mathsf{use}$ & $\mathsf{average}$	\\\tabucline[1.25pt]{-}
				SlowFast R50~\cite{feichtenhofer2019slowfast} & K400~\cite{kay2017kinetics} & 63.9 & 46.5 & 49.4 & 53.2  \\
				\hline
				\multirow{2}{*}{SlowFast R101~\cite{feichtenhofer2019slowfast}} & K600~\cite{carreira2018short} & \multirow{2}{*}{61.0} & \multirow{2}{*}{57.3}  & \multirow{2}{*}{54.0} & \multirow{2}{*}{57.4}\\
				&+AVA2.2~\cite{li2020ava}&&&&\\
				\hline
				X3D-L~\cite{feichtenhofer2020x3d}	&K400~\cite{kay2017kinetics}	&69.2	&49.4	&56.4	&58.3\\\hline
				CLIP~\cite{radford2021learning}	&image-text pairs	&78.5	&62.1	&55.1	&65.2\\\hline
				Ours & MovieCL30K & \textbf{81.5} & \textbf{70.2} & \textbf{61.8} & \textbf{71.1}\\\hline
			\end{tabu}
		}
	\end{center}
	\vspace{-0.4cm}\caption{\small{Comparison of our approach with state-of-the-art pre-trained models on $\mathsf{MCD}$ dataset.}}
	\vspace{-0.4cm}
	\label{tab:cd_result}
\end{table}
\setlength{\tabcolsep}{4pt}

\vspace{0.0cm}\noindent \textbf{a. Mature Content Dataset ($\mathsf{MCD}$):}
\label{ss:data_deep_dive}
%Our dataset focuses on three age-appropriate activities, \textit{i.e.},: (a) $\mathsf{sex}$, (b) $\mathsf{violence}$, and (c) $\mathsf{drug}$-$\mathsf{use}$
%see Figure \ref{eg_cd} for examples
We constructed the $\mathsf{MCD}$ dataset by using Amazon SageMaker Ground Truth as our labeling tool.
%To have consistent and reliable annotations among data-annotators,
We provided annotation guidelines to annotators consisting of $17$ instructions for $\mathsf{sex}$, $34$ for $\mathsf{violence}$ and $14$ for $\mathsf{drug}$-$\mathsf{use}$. Examples of these instructions include: ``intentional murder and/or suicide that involves bloody injury" for $\mathsf{violence}$, and ``strip-tease or erotic dancing with full nudity" for $\mathsf{sex}$.

All our annotators were fluent English speakers which enabled them to understand their provided instructions. All annotators went through multiple training sessions to prepare for their assignment. Each annotator was asked to label sample-videos and received feedback on their labeling quality before they could start to label independently.
Additionally, 20\% of the labeled samples were randomly selected and taken through a mandatory review to ensure quality.

$\mathsf{MCD}$ contains $44,581$ clips from $18,330$ movies and TV episodes with $4,580$, $8,248$, $8,271$, and $23,482$ clips containing $\mathsf{sex}$,  $\mathsf{violence}$, $\mathsf{drug}$-$\mathsf{use}$, and none-of-the-above activities respectively. These clips have a mean duration of $5$ seconds, and were divided into training, validation, and testing sets with 70\%, 10\%, and 20\% ratio respectively.

\vspace{0.0cm}\noindent \textbf{b. Results:}
\noindent We compare our representation with the ones from state-of-the-art models~\cite{feichtenhofer2019slowfast}~\cite{Fan_2021_ICCV}~\cite{feichtenhofer2020x3d}~\cite{radford2021learning} pre-trained on action recognition~\cite{kay2017kinetics}~\cite{carreira2018short}~\cite{gu2018ava} as well as image classification datasets~\cite{radford2021learning}. Specifically, we extract embeddings of $\mathsf{MCD}$ using different pre-trained encoders and use them as input to train a $4$-class MLP for predicting age-appropriate activities.
%We compare our representation with the representations learned using state-of-the-art models on existing action recognition~\cite{feichtenhofer2019slowfast}~\cite{Fan_2021_ICCV}~\cite{feichtenhofer2020x3d} as well as image classification datasets. Specifically, we extract embeddings from action recognition models~\cite{feichtenhofer2019slowfast}~\cite{Fan_2021_ICCV}~\cite{feichtenhofer2020x3d} trained on Kinetics~\cite{kay2017kinetics,carreira2018short} and AVA~\cite{gu2018ava} as well as CLIP visual encoder~\cite{radford2021learning} on clips in our dataset, extract the embeddings from these models, and provide them as inputs to the aforementioned MLP model for training.
Table~\ref{tab:cd_result} presents the AP results on our test data, and shows that our scene representation outperforms the alternatives by a large margin. Notice that CLIP visual feature performed close to our representation on the task of $\mathsf{sex}$ but not other tasks. This might indicate that in the pre-training data of CLIP, there might be a considerable amount of image-text pairs related to $\mathsf{sex}$ concepts. Similar observations~\cite{birhane2021multimodal} have been made on other large-scale datasets~\cite{schuhmann2021laion}.
\vspace{-0.0cm}
\section{Conclusions and Future Work}
\label{conclusion}

\noindent We presented a novel contrastive learning approach that uses movie metadata to learn a general-purpose scene-level representation. Our approach adopts recent developments in vision transformer, where by incorporating the interpolation of positional embedding, we were able to train a scene encoder that can take variable-length multi-shot inputs. We empirically demonstrated the effectiveness of our approach using different types of movie metadata including co-watch, genre, synopsis. Results on a variety of tasks from multiple benchmark datasets highlight the strengths of our approach. We also presented a new application of our scene representation for video moderation focused on three age-appropriate activities, where our approach outperformed state-of-the-art action recognition features. 

Going forward, we will further improve the efficiency of our approach to help it scale even more. We will also explore other types of movie metadata to find a representation effective for additional downstream tasks. Furthermore, there are additional long-form content (\textit{e.g}, music, news articles, books) where acquiring fine-grained labels is costly. We will apply our approach on these long-form content to enable downstream tasks from multiple domains.

%%%%%%%%% REFERENCES
{\small
\bibliographystyle{ieee_fullname}
\bibliography{ref}
}

\end{document}

\title{Supplementary Material\\Movies2Scenes: Using Movie Metadata to Learn Scene Representation}
%Learning Scene Representations Using Movie Metadata 
\author{Shixing Chen \quad Chun-Hao Liu \quad Xiang Hao \quad Xiaohan Nie \quad Maxim Arap \quad Raffay Hamid\\
Amazon Prime Video\\
{\tt\small \{shixic, chunhaol, xianghao, nxiaohan, maxarap, raffay\}@amazon.com}
%\and
}
\maketitle

\noindent In this supplementary material, we provide two sections to better support the arguments and results in the main paper: (a) more details of the specific settings in our experiments for better reproducibility, and (b) more extensive qualitative results to better demonstrate the interpretability of our learned representations. The sections will be presented with information corresponding to different datasets used in the main paper including: Movie Contrastive Learning 30K ($\mathsf{MovieCL30K}$) dataset, Long-Form Video Understanding (LVU) dataset~\cite{Wu_2021_CVPR}, MovieNet dataset~\cite{huang2020movienet}~\cite{rao2020local} and Mature Content Dataset ($\mathsf{MCD}$).

\section{Experiment Details}

\noindent We use PyTorch $1.8$~\cite{paszke2019pytorch} as our deep learning library and NVIDIA Tesla A100/V100 GPUs for computation. During contrastive learning, we use $8$ GPUs with distributed data and model parallelism. For supervised learning of MLP on downstream tasks, only $1$ GPU is needed.

\subsection{MovieCL30K}
\noindent This section corresponds to $\S$$4.1.2$ in the main paper.

\noindent \textbf{a. Shot-encoder:} Our shot encoder has two key differences comparing with ShotCoL~\cite{Chen_2021_CVPR}. First, we can select the positive keys during training, which makes training more efficient where we do not use the stale positive keys for epochs before updating them.
%Second, ShotCoL~\cite{Chen_2021_CVPR} focused on learning a representation that is most useful for the scene boundary detection task, so it could benefit from contextual and semantic information in a neighborhood size similar to the length of a scene. That is, given a query shot, the positive key was selected to be the most similar shot in a neighborhood of $16$ shots~\cite{Chen_2021_CVPR}, so that when similar shots appear in a sequence of shots, it can be used as the cue for scene boundary detection. In our case, we still use the sequential relationship in movies but limit the neighborhood size.
Second, ShotCoL~\cite{Chen_2021_CVPR} focused on learning a representation that is most useful for the scene boundary detection task, so it could benefit from contextual and semantic information in a neighborhood size similar to the length of a scene. However, we observed that when the neighborhood size is relatively large (e.g., 16) as selected in ShotCoL, the positive key may end up being almost identical to the query. This is still useful information for scene boundary detection task because there could be almost identical shots in a scene. However, for our objective to learn a representation that focuses on appearance, this may reduce the effectiveness of representations because the positive key is more similar to augmented images which were demonstrated to be less effective in~\cite{Chen_2021_CVPR}.

\noindent \textbf{b. Movie-level similarity learning:} When using movie metadata to train the movie encoder (Figure 2 in the main paper) on $\mathsf{MovieCL30K}$, we used SGD to optimize with a learning rate of $0.1$, batch size of $256$ and epoch number of $100$. The same set of hyper-parameters was applied to all three types of movie metadata (co-watch, genre, and synopsis). Recall that within each batch, there are $256$ pairs of movies represented by feature matrices extracted from our shot encoder, and the dimension of each feature matrix is $1024$$\times$$512$. These pairs are passed through $\mathbf{E}_{\textrm{movie}}$ to predict whether two movies are similar based on movie metadata.

\iffalse
\begin{figure}[t]
	\centering
	\vspace{-0.1cm}
	\includegraphics[width=0.5\textwidth]{acc_curves_6x.png}
	\vspace{-0.3cm}
	\caption{\small{During pre-training on different movie-level metadata, top-1 and top-5 accuracy results are presented. By using co-watch metadata, the accuracy is higher than other information. This may indicate that the representations learned on co-watch are more discriminating, which leads to better performance on downstream tasks shown in Table 1 of the main paper.}}
	\vspace{-0.1cm}
	\label{acc_curves}
\end{figure}
\fi

\setlength{\tabcolsep}{4pt}
\begin{table*}[t]
	\begin{center}
		\small{
			\begin{tabu}{c|ccccccc|cc}
				\hline
				\multirow{2}{*}{Hyper-parameters}	&\multicolumn{7}{c|}{Classification}	&\multicolumn{2}{c}{Regression}\\
				&place	&director	&relation	&speak	&writer	&year	&genre	&view	&like\\\tabucline[1.25pt]{-}
				learning rate		&0.1	&1.0	&0.01	&0.1	&0.5	&0.1	&0.1	&0.01	&0.1\\
				batch size	&128	&256	&16	&256	&32	&128	&256	&16	&16\\
				epoch	&400	&400	&400	&400	&400	&400	&400	&500	&500\\
				dropout	&0.1	&0.5	&0.5	&0.1	&0.25	&0.75	&0.1	&0.8	&0.5\\\hline
			\end{tabu}
		}
	\end{center}
	\vspace{-0.2cm}\caption{\small{Hyper-parameters used in the MLPs for LVU tasks.}}\vspace{-0.0cm}
	\label{lvu_hyp}
\end{table*}
\setlength{\tabcolsep}{4pt}

\begin{figure*}[!ht]
	\centering
	\vspace{-0.1cm}
	\includegraphics[width=0.9\textwidth]{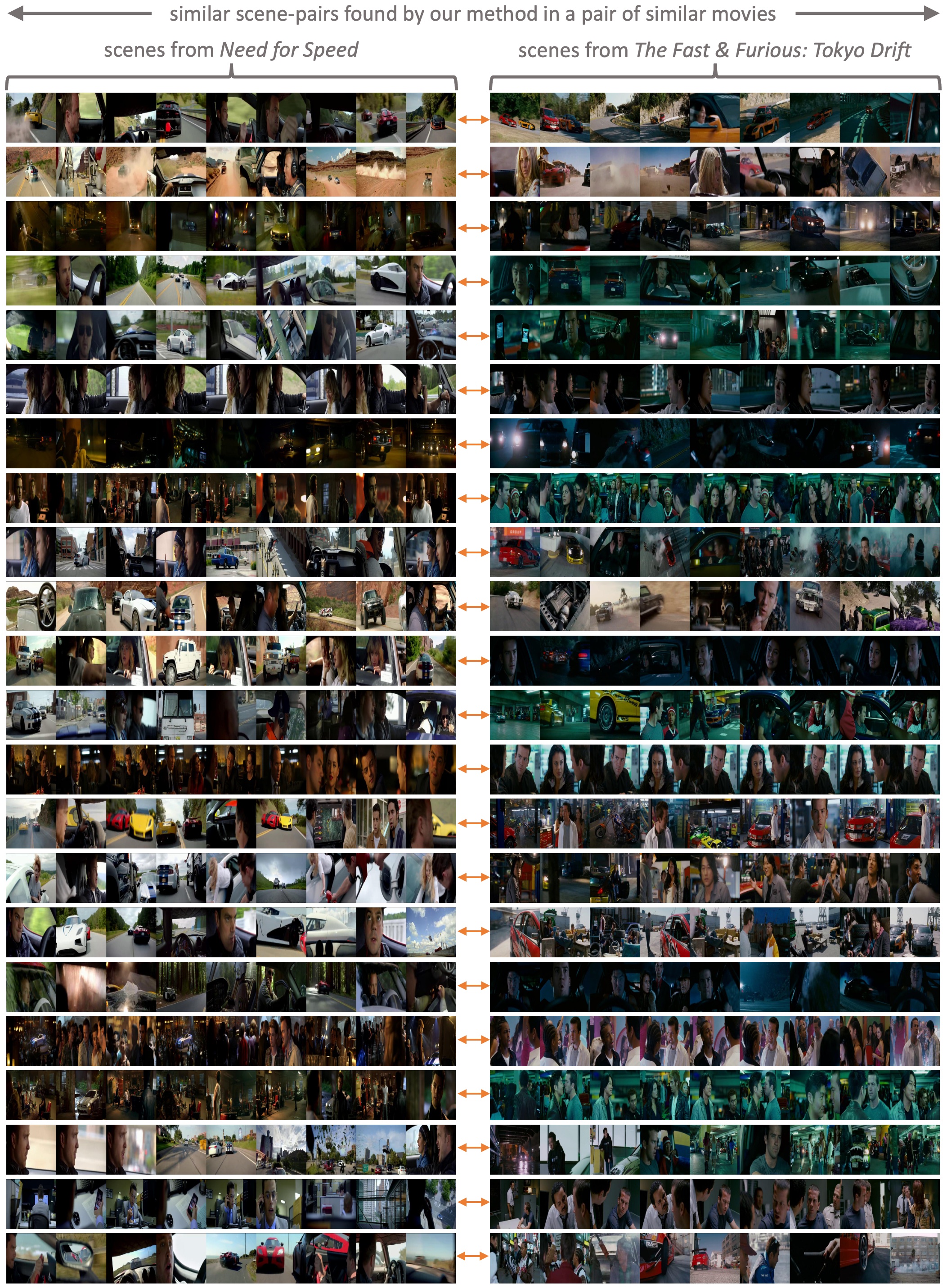}
	\vspace{-0.3cm}
	\caption{\small{\textbf{Similar scene-pairs found by our representation - }Given a similar movie-pair \textit{Need For Speed} and \textit{The Fast \& Furious: Tokyo Drift}, we present representative examples from the set of similar scene-pairs (connected by orange arrow) found by our representation (sorted by similarity). Comparing with the ones found by CLIP visual feature~\cite{radford2021learning} in Figure~\ref{clip_scene} and Merlot Reserve feature~\cite{zellers2022merlotreserve} in Figure~\ref{merlot_scene}, these scene-pairs are more thematically meaningful, which contributes to the effectiveness of our representation on downstream tasks related to semantic scene understanding.}}
	\vspace{-1.0cm}
	\label{sims_scene}
\end{figure*}

\begin{figure*}[!ht]
	\centering
	\vspace{-0.1cm}
	\includegraphics[width=0.9\textwidth]{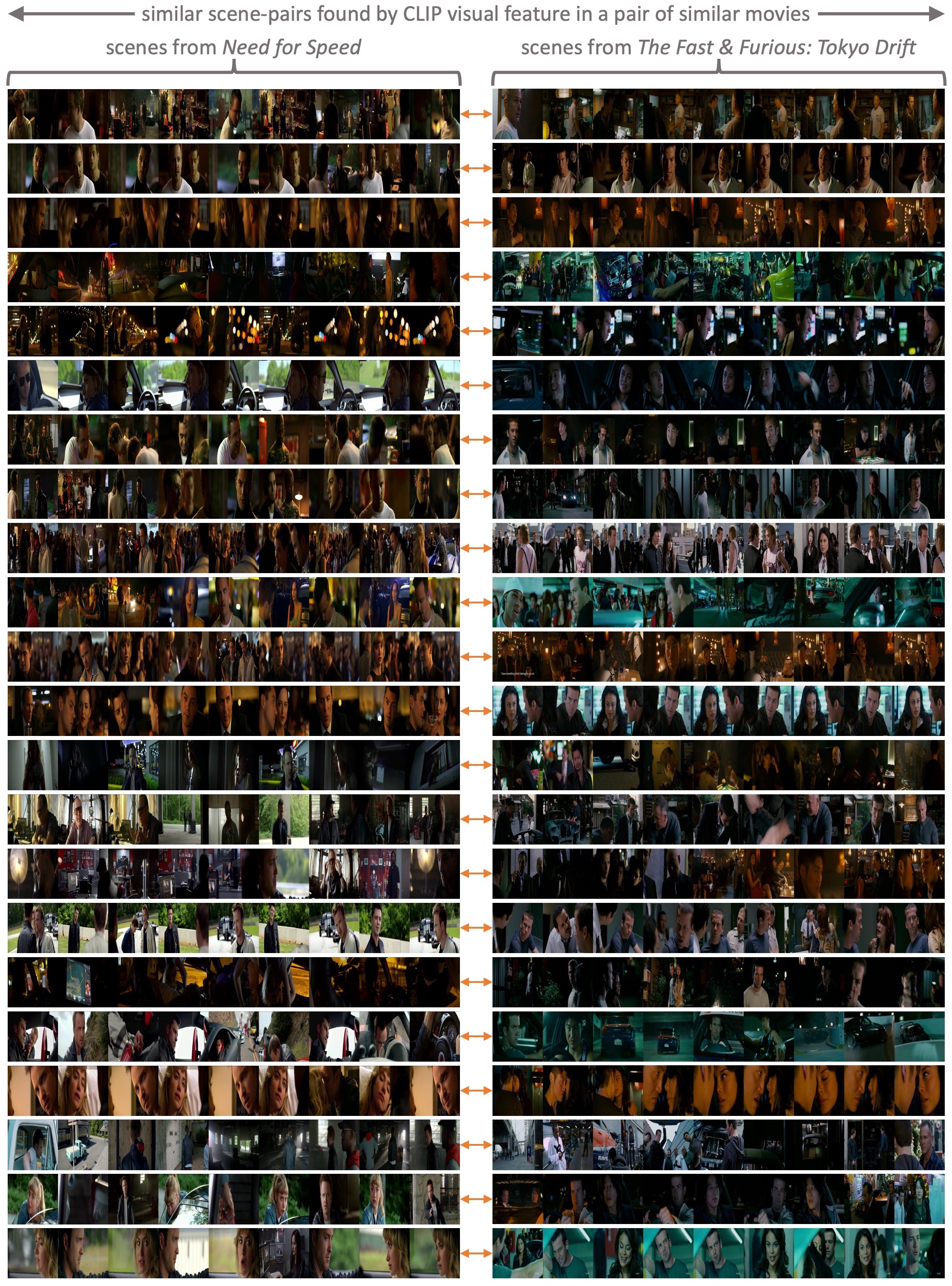}
	\vspace{-0.3cm}
	\caption{\small{\textbf{Similar scene-pairs found by CLIP - }Given a similar movie-pair \textit{Need For Speed} and \textit{The Fast \& Furious: Tokyo Drift}, we present representative examples from the set of similar scene-pairs (connected by orange arrow) found by CLIP visual feature~\cite{radford2021learning} (sorted by similarity). Comparing with the ones found by our representation in Figure~\ref{sims_scene}, these scene-pairs focus more on appearance-based similarity and are mostly related to human faces, which are not sufficiently semantically meaningful for semantic scene understanding.}}
	\vspace{-1.0cm}
	\label{clip_scene}
\end{figure*}

\begin{figure*}[!ht]
	\centering
	\vspace{-0.1cm}
	\includegraphics[width=0.9\textwidth]{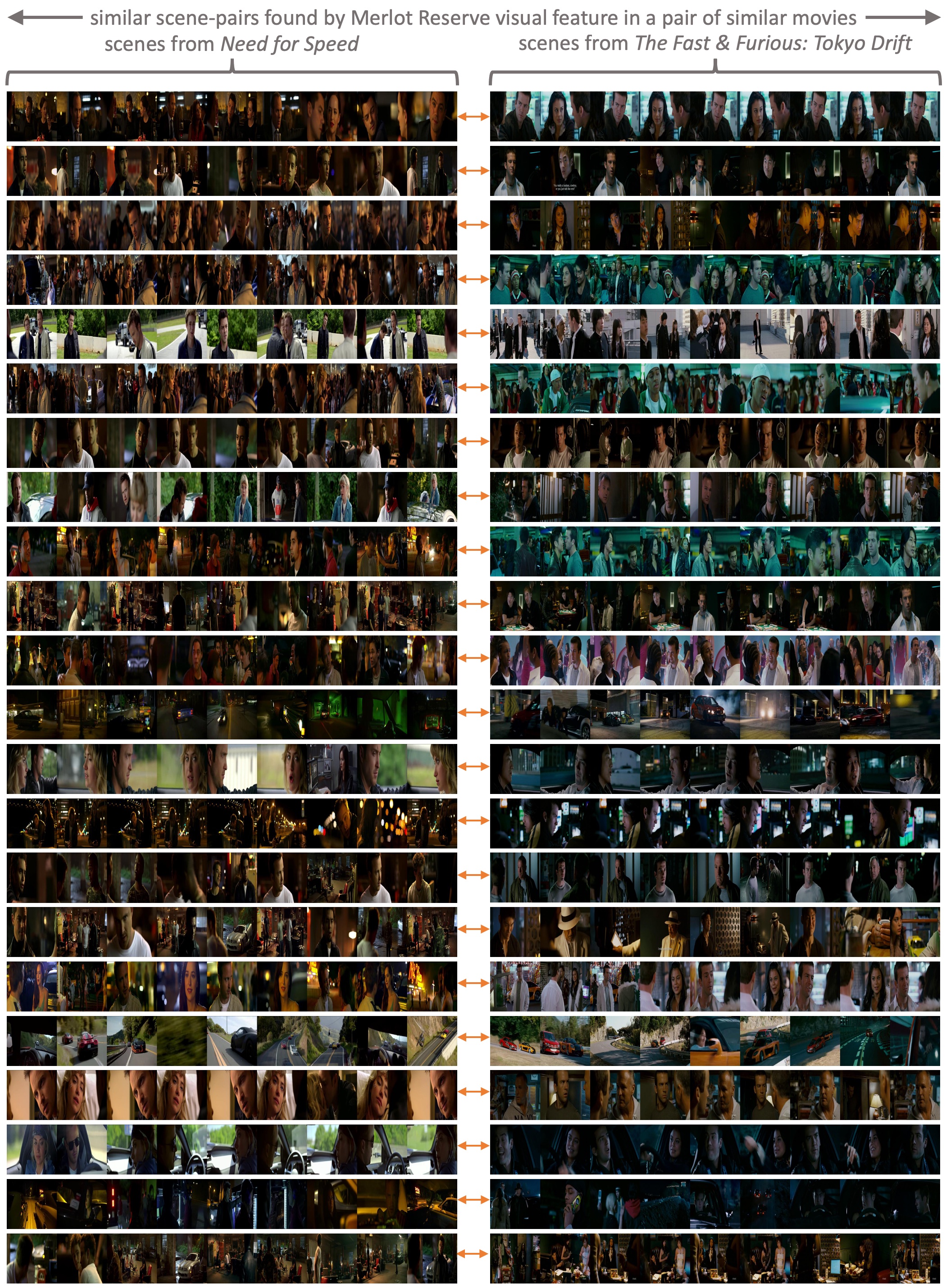}
	\vspace{-0.3cm}
	\caption{\small{\textbf{Similar scene-pairs found by Merlot Reserve - }Given a similar movie-pair \textit{Need For Speed} and \textit{The Fast \& Furious: Tokyo Drift}, we present representative examples from the set of similar scene-pairs (connected by orange arrow) found by Merlot Reserve visual feature~\cite{zellers2022merlotreserve} (sorted by similarity). Comparing with the ones found by our representation in Figure~\ref{sims_scene}, these scene pairs are more similar to the ones found by CLIP in Figure~\ref{clip_scene}, which focus more on appearance-based similarity and are mostly related to human faces, which are not sufficiently semantically meaningful for semantic scene understanding.}}
	\vspace{-1.0cm}
	\label{merlot_scene}
\end{figure*}

\noindent \textbf{c. Scene contrastive learning:} After movie-level similarity learning, we select the set of similar scene-pairs based on the learned space in $\mathbf{E}_{\textrm{movie}}$. Specifically, after extracting features of all shots in each input movie by $\mathbf{E}_{\textrm{shot}}$ , the movie is represented as a matrix of \textrm{M}$\times$$512$, where \textrm{M} is the number of shots in the movie. Notice that the length of the movie is no longer restricted to $1024$, so that all shots can be considered during similar scene selection. The feature matrix is then passed to $\mathbf{E}_{\textrm{movie}}$ before the last fully-connected layer, and becomes a new feature matrix. Similar process is done on another movie considered similar to the input movie with \textrm{N} shots, and the shot adjacency matrix $\mathbf{A}$ of these two movies takes the size of \textrm{M}$\times$\textrm{N}. We then go through all $9$$\times$$9$ windows in $\mathbf{A}$ with stride $1$, and calculate the average value in each window to represent the scene-level similarity of the two movies. Finally, we pop the scene-pairs with top $50\%$ highest scene-level similarity scores while keeping the selected scenes to be non-overlapping. This generates a set of scene-pairs for each type of movie metadata.

With the generated set of scene pairs, we use them for scene contrastive learning following the MoCo framework~\cite{He_2020_CVPR}, while substituting encoder to be ViT~\cite{dosovitskiy2020vit} and optimization to be AdamW~\cite{loshchilov2019decoupled} instead of SGD~\cite{goyal2017accurate}.
Specifically, following~\cite{He_2020_CVPR}, we use feature dimension of 128, queue size of $65,536$, MoCo momentum of 0.999 and softmax temperature of 0.07 during momentum contrastive learning. Following~\cite{dosovitskiy2020vit}\cite{chen2021mocov3}, we use learning rate of $1.5$\textrm{e}$-$$4$, weight decay of $0.01$, number of warm-up epochs of $40$, batch size of $128$ and number of epoch of $100$.

%We present the top-1 and top-5 accuracy during pre-training on different movie metadata (co-watch, genre and synopsis) in Figure~\ref{acc_curves}. We can see that co-watch gets the highest top-1 and top-5 accuracy during pre-training, while synopsis gets the lowest ones, which indicates that more discriminating features have been learned from co-watch leading to give better performance on downstream tasks validated in Table 1 in the main paper. Also notice that the accuracy at the beginning of each epoch is less stable therefore producing the peaks in Figure~\ref{acc_curves}.

\subsection{LVU}
%\noindent This section corresponds to $\S$$4.2$ in the main paper.
\noindent When producing the results in Table 2 in the main paper, following~\cite{Wu_2021_CVPR} where the model of each task is trained separately with parameters selected by validation set, we selected the parameters and hyper-parameters of MLP for each task by the validation set in LVU and presented corresponding results on test set. The hyper-parameters used in MLPs of each task on representations pre-trained by co-watch is shown in Table~\ref{lvu_hyp}.

\subsection{MovieNet}
\noindent This section corresponds to $\S$$4.3$ in the main paper.

\noindent \textbf{a. Place tagging:} For the results of Ours in Table 3 in the main paper, we used MLP with two 512-dimensional hidden layers optimized by SGD with leaning rate of $5.0$, dropout of $0.25$, epoch number of $200$ and batch size of $512$. The problem was formulated as a multi-label classification task and optimized by BCEWithLogitsLoss~\cite{paszke2019pytorch}.

\noindent \textbf{b. Scene boundary detection:} For the results of Ours in Table 4 in the main paper, we used MLP with two 512-dimensional hidden layers optimized by SGD with leaning rate of $0.03$, dropout of $0.8$, epoch number of $800$ and batch size of $4096$.

\subsection{MCD}

\noindent We used SlowFast 8x8 R50 and SlowFast 8x8 R101 for the SlowFast models~\cite{feichtenhofer2019slowfast} used in Table 5 in the main paper. Both models take 64 frames from each video clips. When extracting representation from the SlowFast models, we used the average pooling layers before the final classification layer, and the representation has 2304 dimensions concatenated from the slow and fast pathways. Similarly for the X3D-L model~\cite{feichtenhofer2020x3d}, we used the fully connected layer before the final classification layer, which has 2048 dimensions, and the X3D-L model takes 16 frames from each video clips. For the CLIP model\cite{radford2021learning}, we used ViT-B/16 based visual encoder, and it takes the same 9 frames as the input to our model from each video clip. We first extracted the embeddings with 512 dimensions from each frame and then do an average pooling across all 9 frames, which is used as the representation for the CLIP model. For training the model to classify the age-appropriate activities, we used a 3-layer MLP model with 512 nodes in the hidden layers.

\section{Additional Results}

\begin{table*}[t]
	\begin{center}
		\smaller{
			\begin{tabu}{c!{\vrule width 1.25pt}ccc|ccc|ccc}
				\hline
				Models	&\multicolumn{3}{c}{CLIP~\cite{radford2021learning}}&	\multicolumn{3}{c}{SlowFast~\cite{feichtenhofer2019slowfast}}&	\multicolumn{3}{c}{Ours}\\\hline
				Architecture&\multicolumn{3}{c}{ViT-B/16~\cite{dosovitskiy2020vit}}&	\multicolumn{3}{c}{ResNet-101~\cite{he2016deep}}	&\multicolumn{3}{c}{ViT-B/16~\cite{dosovitskiy2020vit}}\\
				Pre-training data	&\multicolumn{3}{c}{400M image-text pairs}	&\multicolumn{3}{c}{Kinetics~\cite{carreira2018short}+AVA~\cite{li2020ava}}&\multicolumn{3}{c}{MovieCL30K}\\
				Pre-training task	&\multicolumn{3}{c}{image-text similarity}	&\multicolumn{3}{c}{action recognition}&\multicolumn{3}{c}{scene contrast}\\\tabucline[1.25pt]{-}
				%\multirow{2}{*}{Place}	&\multicolumn{9}{c}{Precision (\%)}\\\cline{2-10}
				&top-1	&top-5	&top-10	&top-1	&top-5	&top-10	&top-1	&top-5	&top-10	\\\hline
				office	&30	&20	&19	&0	&16	&17	&30	&36	&26	\\
				airport	&0	&15	&10	&0	&0	&10	&0	&15	&10	\\
				school	&28.57	&26.66	&25	&76.19	&20.47	&32.14	&50	&49.04	&45.95	\\
				hotel	&35.71	&28.57	&29.28	&0	&0	&10	&42.85	&20	&24.28	\\
				prison	&52.94	&42.94	&42.94	&35.29	&40	&20.29	&58.82	&45.88	&40	\\
				restaurant	&20	&13.99	&13.99	&0	&20	&10	&40	&16	&14	\\ \hline
				$\mathsf{all}$ $\mathsf{queries}$	&35.08	&29.64	&28.85	&38.59	&22.63	&21.84	&\textbf{47.36} &\textbf{39.29}	&\textbf{35.70}	\\\hline
			\end{tabu}
		}
	\end{center}
	\vspace{-0.5cm}\caption{\small{The place-labeled scenes in LVU data~\cite{Wu_2021_CVPR} are formulated to a retrieval setting, where the goal is to retrieve similar scenes from training-set given query scenes from validation-set. Representations pre-trained on different configurations as specified in the table are compared. The precision results for different place categories are reported with the size of retrieved set to be $1$, $5$ and $10$.}}\vspace{-0.0cm}
	\label{res_retrieval}
\end{table*}

\begin{figure*}[!htb]
	\centering
	\includegraphics[width=0.9\textwidth]{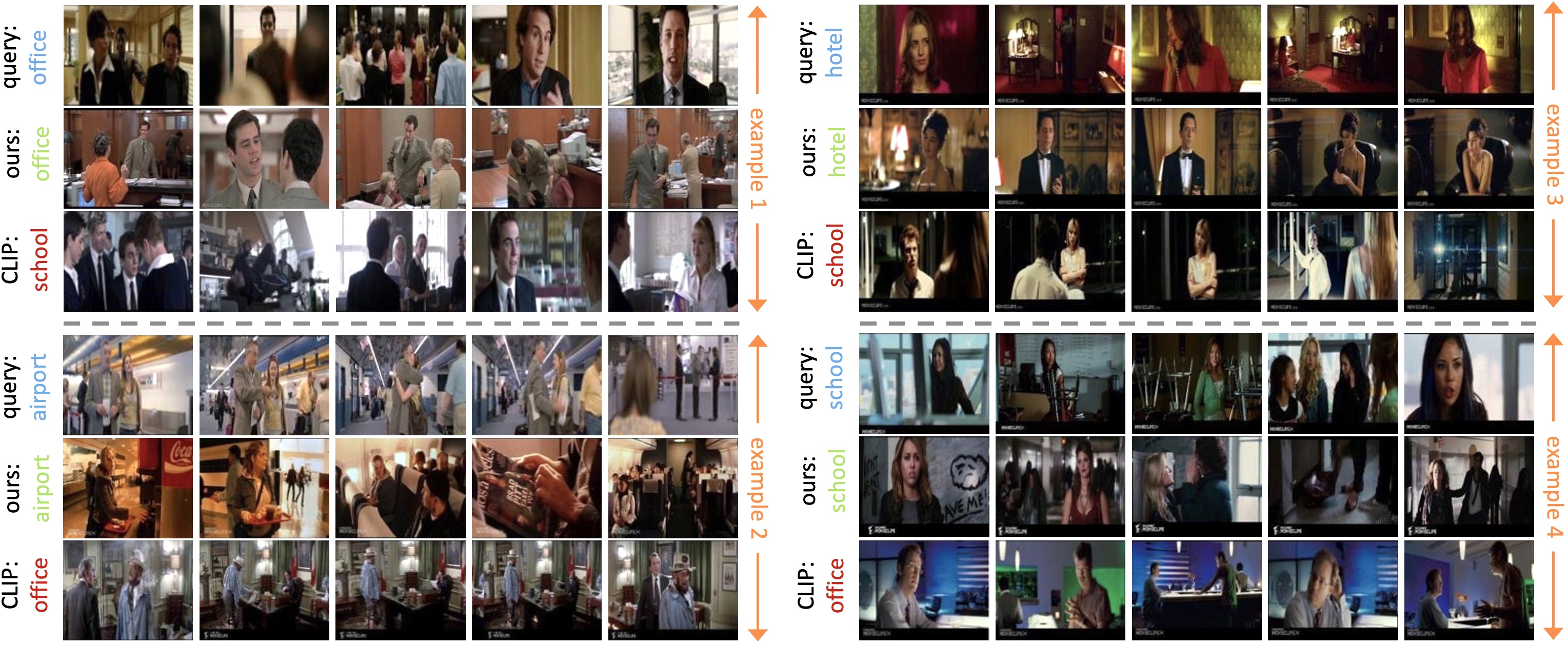}
	\vspace{-0.2cm}\caption{\small{Qualitative results of place retrieval using LVU data~\cite{Wu_2021_CVPR} are shown. For each query scene in validation-set, two similar scenes from training-set are retrieved based on ours and CLIP visual representation~\cite{radford2021learning}. Example results show that our feature can capture both scene-appearance as well as their broader thematic signature, while CLIP~\cite{radford2021learning} can only capture scene-appearance effectively.}}\vspace{-0.1cm}
	\label{fig_retrieval}
\end{figure*}

\subsection{MovieCL30K}

\noindent We present the similar scene pairs selected by our scene representation learned on co-watch in Figure~\ref{sims_scene}. We also present the similar scene-pairs found by pre-trained CLIP visual features~\cite{radford2021learning} in Figure~\ref{clip_scene} and the ones found by pre-trained Merlot Reserve visual features~\cite{zellers2022merlotreserve} in Figure~\ref{merlot_scene}, respectively. We can see that scene-pairs found by our approach are significantly more \textbf{thematically similar}, while the ones found by other features focus much more on \textbf{appearance similarity}. Moreover, CLIP~\cite{radford2021learning} and Merlot Reserve~\cite{zellers2022merlotreserve} features produce results that are mostly related to human faces, which is not sufficiently useful for general-purpose semantic scene understanding. These observations provide insights about the effectiveness of our approach on a wide variety of downstream tasks related to semantic scene understanding compared to other state-of-the-art representations. Lastly, we can also noticed that the scenes pairs found by Merlot Reserve visual features~\cite{zellers2022merlotreserve} are similar to the ones found by CLIP visual features~\cite{radford2021learning}, which can indicate that the added audio modality in Merlot Reserve in not directly influencing the distribution of embedding in visual encoder before the modalities are fused.
%and has been validated by the qualitative results throughout the experiment section in the main paper.

\begin{figure}[!t]
	\centering
	\vspace{-0.1cm}
	\includegraphics[width=0.475\textwidth]{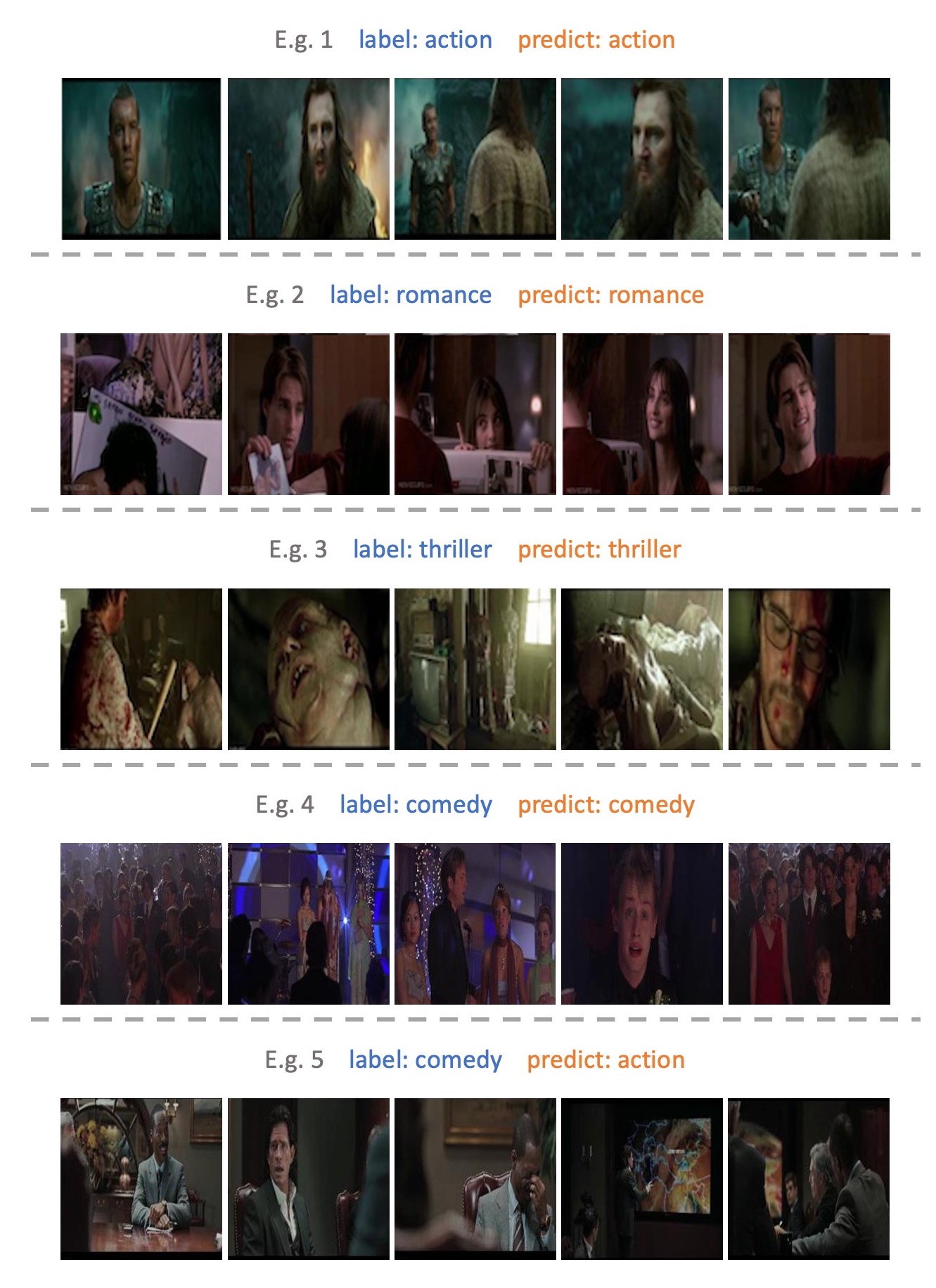}
	\vspace{-0.5cm}
	\caption{\small{Additional qualitative results on LVU datasets~\cite{Wu_2021_CVPR}. Five example scenes with their ground truth genre labels as well as our predictions are shown. For example 5, our prediction is different from the label even though looking only at the visual content of this particular scene it makes sense to infer it as a romantic scene. Our hypothesis is that as the genre label of the LVU dataset was acquired from movie-level meta data, sometimes it is not directly applicable to the genre of all of the constituent scenes of a movie.}}
	\vspace{-0.5cm}
	\label{qua_lvu_genre}
\end{figure}

\begin{figure}[!t]
	\centering
	\vspace{-0.5cm}
	\includegraphics[width=0.475\textwidth]{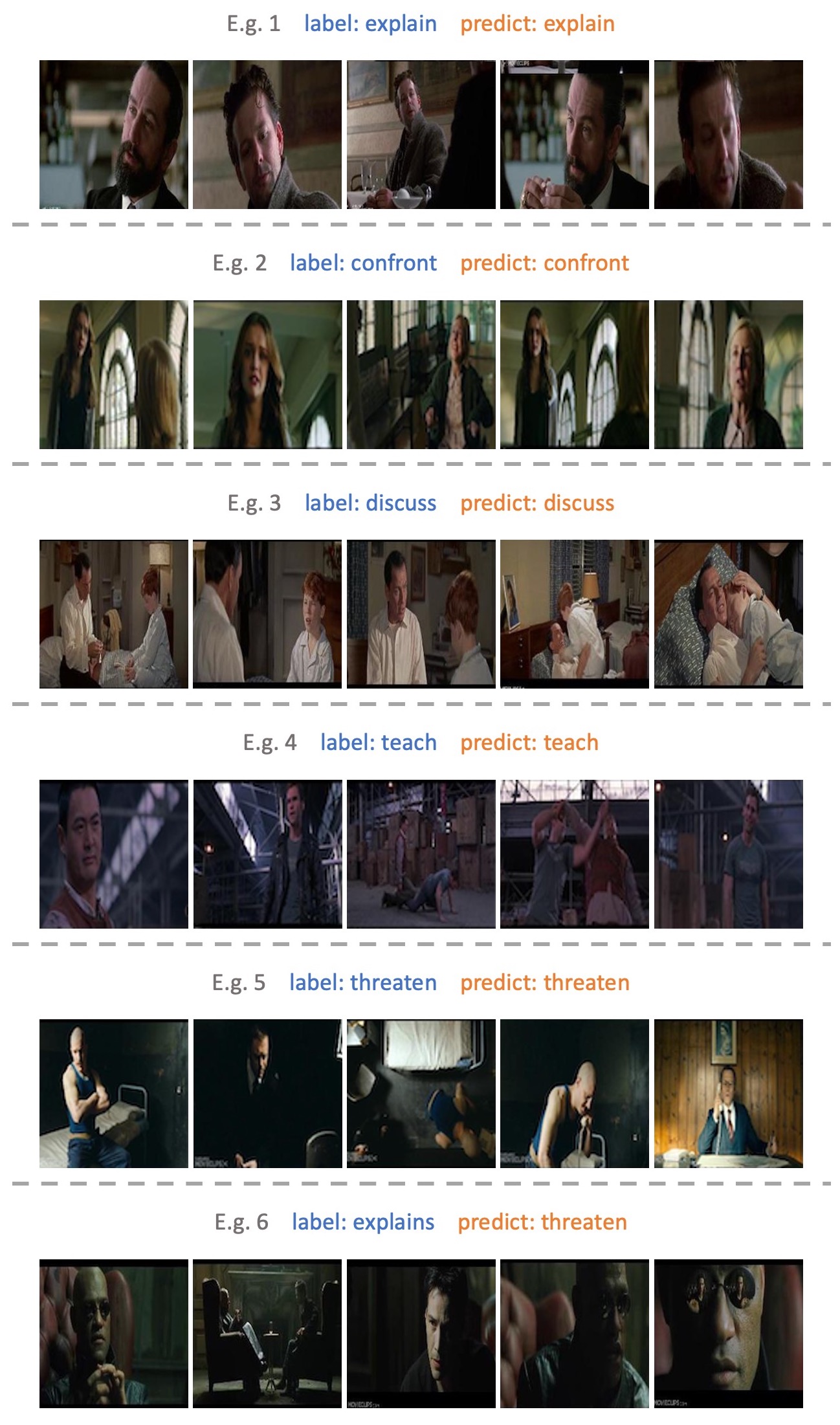}
	\vspace{-0.5cm}
	\caption{\small{Six examples from the way-of-speaking prediction task in LVU~\cite{Wu_2021_CVPR}. Results show that when visual information is sufficient to predict way-of-speaking, our method can perform well, and for cases like example 6, it might be beneficial to add audio modality to further improve the accuracy.}}
	\vspace{-0.3cm}
	\label{qua_lvu_speak}
\end{figure}

\begin{figure}[!t]
	\centering
	\vspace{-0.0cm}
	\includegraphics[width=0.45\textwidth]{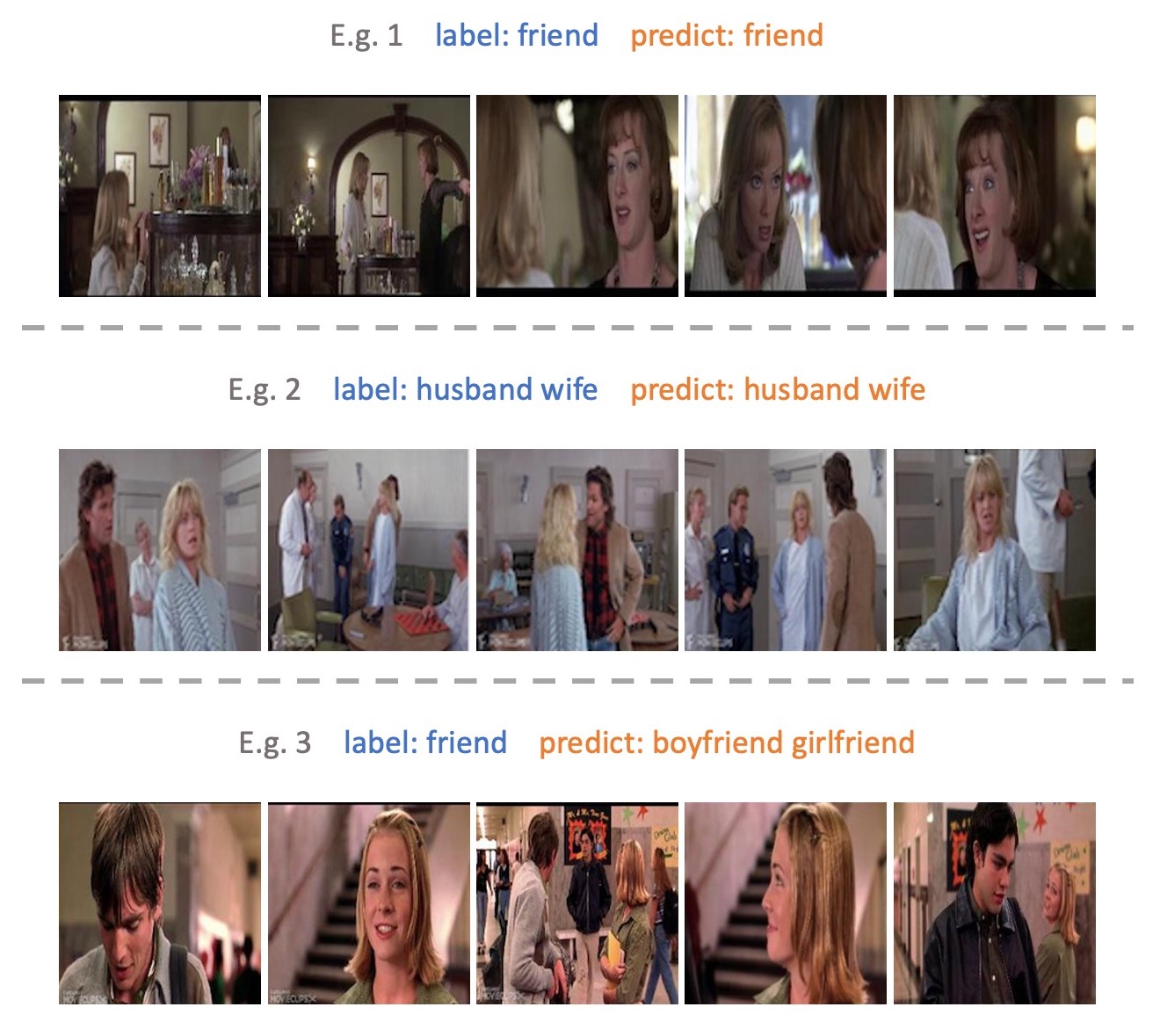}
	\vspace{-0.3cm}
	\caption{\small{Examples of relationship prediction task in LVU~\cite{Wu_2021_CVPR}. Sometimes the relationship can be ambiguous when there are more than two leading characters in the scene. Also, the uncertainty of some relationships makes prediction even more challenging. For example, it can be difficult to distinguish husband wife and boyfriend girlfriend without semantically understanding of context and plot.}}
	\vspace{-0.0cm}
	\label{qua_lvu_relation}
\end{figure}

\begin{figure}[!t]
	\centering
	\vspace{-0.0cm}
	\includegraphics[width=0.49\textwidth]{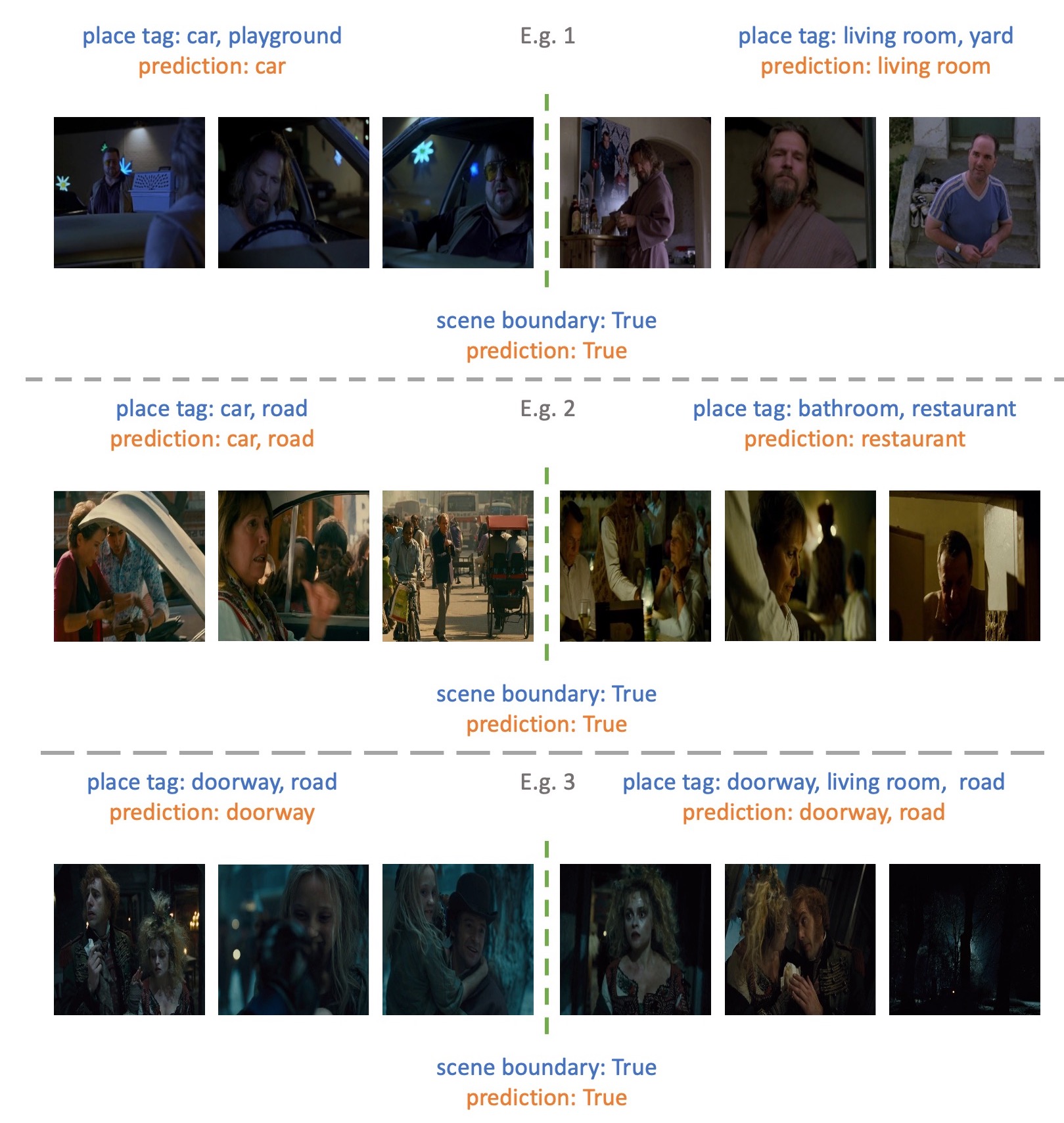}
	\vspace{-0.5cm}
	\caption{\small{Qualitative results on MovieNet place tagging~\cite{huang2020movienet} and scene boundary detection tasks~\cite{rao2020local}. Although our model outperformed existing state-of-the-art models by a large margin in Table 3 in the main paper, multi-labeled place tagging is still a really challenging problem. Some tags may not be apparent and the intra-category variance is large in this task.}}
	\vspace{0.0cm}
	\label{qua_mn}
\end{figure}

\begin{figure}[!t]
	\centering
	\includegraphics[width=0.475\textwidth]{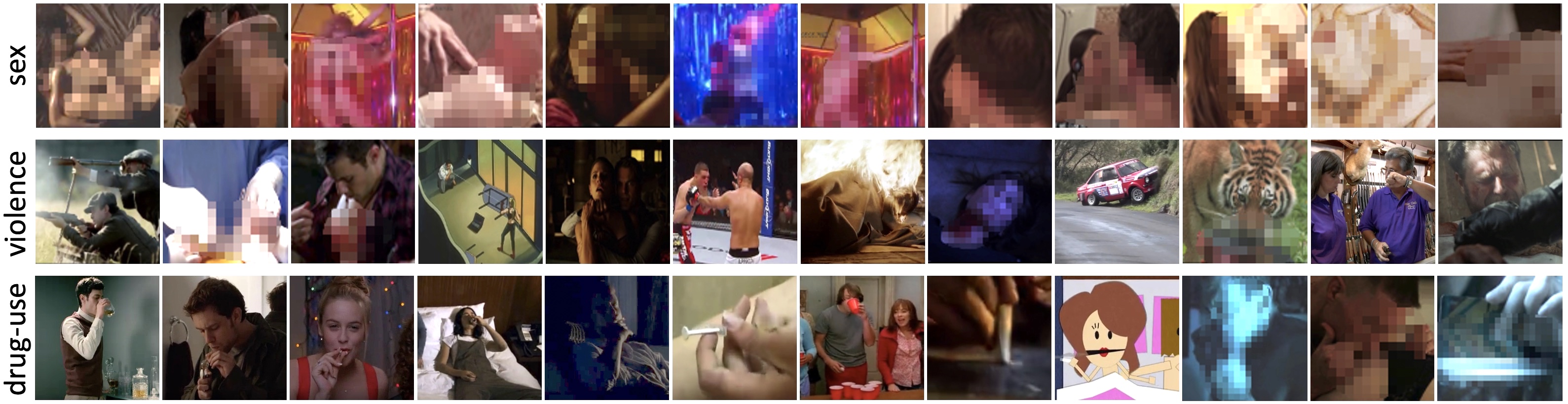}\vspace{-0.2cm}
	\caption{\small{Examples of $3$ types of age-appropriate activities in our data. Sensitive parts of images have been intentionally redacted here. See supplementry materials for more examples.}}\vspace{-0.0cm}
	\label{eg_cd}
\end{figure}

\begin{figure}[!t]
	\centering
	\vspace{0.0cm}
	\includegraphics[width=0.475\textwidth]{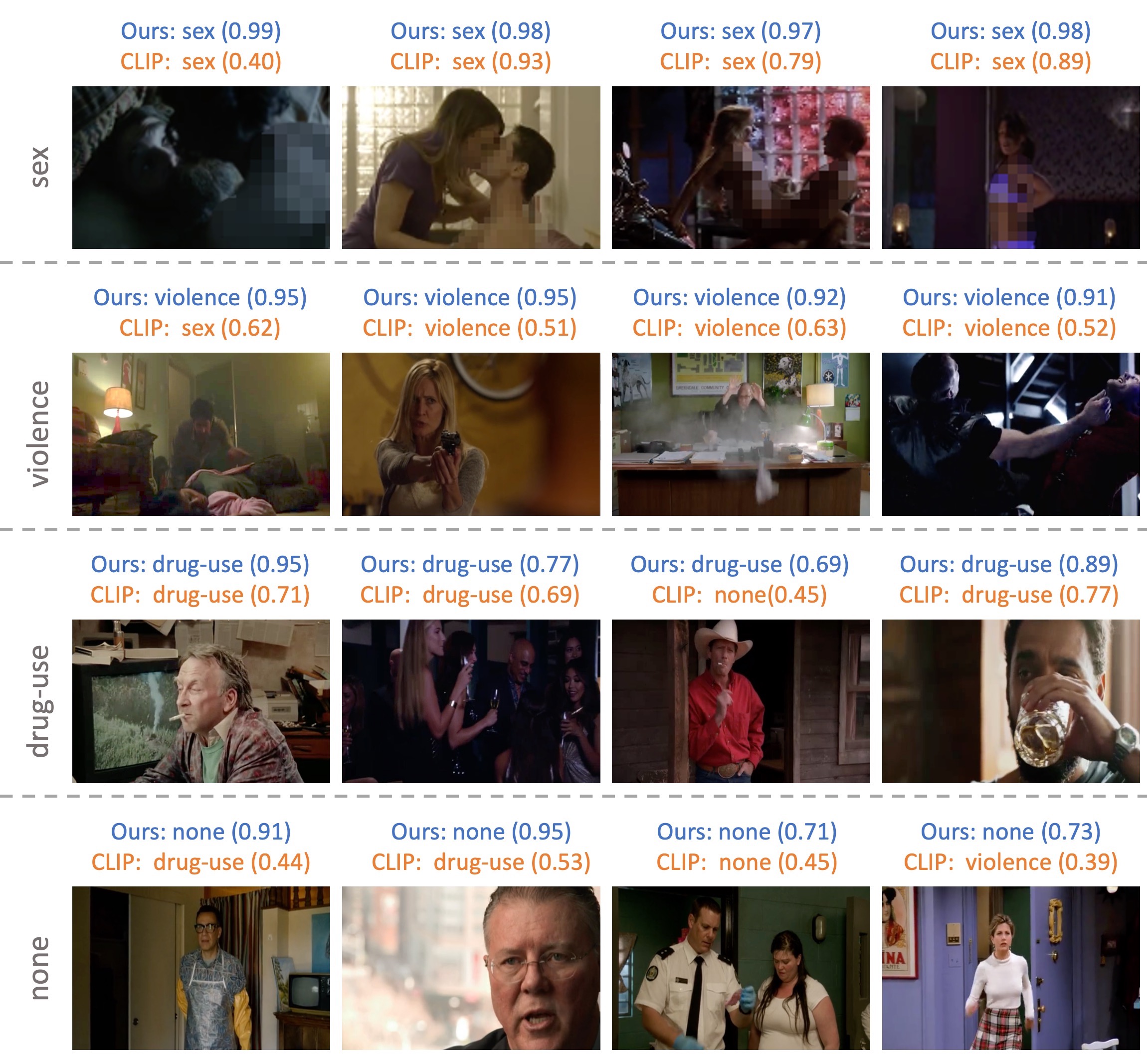}
	\vspace{-0.5cm}
	\caption{\small{Representative examples in $\mathsf{MCD}$ comparing the predictions based on our representation with CLIP visual feature. Sensitive parts of some images are intentionally redacted here.}}
	\vspace{-0.1cm}
	\label{qua_mcd}
\end{figure}

\subsection{LVU}

\noindent To demonstrate the effectiveness of our learned scene-representation, we use the place-labeled scenes from LVU data~\cite{Wu_2021_CVPR} in a retrieval-setting. Specifically, using query scenes each with a particular place-label from the validation-set, we retrieve $1$, $5$, and $10$ nearest neighbors from the training-set using their $\textrm{L}_2$ distances.
Precision results for various settings are given in Table~\ref{res_retrieval} where our encoder is compared with the pre-trained visual encoder of CLIP~\cite{radford2021learning} and the SlowFast model~\cite{feichtenhofer2019slowfast} pre-trained on Kinetics~\cite{carreira2018short} and AVA~\cite{li2020ava}. 

Moreover, to provide qualitative insights into the effectiveness of our learned scene-representation, Figure~\ref{fig_retrieval} shows the retrieval results using an example query for $4$ of the $6$ categories based on ours as well as CLIP~\cite{radford2021learning} visual representation. It can be seen that although CLIP visual representation can capture local appearance-based patterns effectively, it is not able to capture longer-duration semantic aspects of scenes. In contrast, our representation is able to capture the appearance as well as semantic aspects of scenes effectively, and is therefore able to avoid the types of confusions that confound the CLIP representation.

%In Table~\ref{res_retrieval}, when comparing with CLIP feature~\cite{radford2021learning}, we adopted the visual encoder from pre-trained CLIP, and extracted features for each shot in the input scene, followed by average pooling.
We applied average pooling instead of concatenation on CLIP feature~\cite{radford2021learning} because during retrieval, we do not want the order of shots to influence the results, thus, having one vector per input scene is reasonable. Notice that for the place of airport, both our representation and CLIP have top-1 accuracy of $0$, it is because the number of query is very limited in the validation set of LVU~\cite{Wu_2021_CVPR} ($4$ airport-labeled scenes), and thus the airport example presented in Figure~\ref{fig_retrieval} comes from the top-2 retrieval result.

In Figure~\ref{qua_lvu_genre}, we show five example scenes with their ground truth genre in LVU as well as our prediction. We can see that our model was able to capture the feature that is useful for correct genre prediction in  most cases. For cases like example 5 on the last row, we assume it is because it is sometimes difficult to identify genre from just one scene in the movie, and since the meta information like genre in LVU was retrieved from IMDB entries~\cite{Wu_2021_CVPR}, they can not always reflect the genre of a specific scene.

In Figure~\ref{qua_lvu_speak}, we show six examples from the way-of-speaking prediction task in LVU. We can see that when the visual information is sufficient to make predictions, our model can perform well, but for cases like example 6 on the last row, it can be insufficient to predict just based on visual cues, and this might indicate that for tasks like way-of-speaking prediction, it may be beneficial to include audio modality for better accuracy. This also corresponds to the results in Table 2 of the main paper, where the accuracy of way-of-speaking is lower than other tasks.
For relationship prediction in Figure~\ref{qua_lvu_relation} we can see that the model can make ambiguous predictions when there are more than one pairs of characters in the scene, and this may indicate that when analyzing relationship in scenes, it can be helpful to focus more on leading characters.

\subsection{MovieNet}

\noindent We present qualitative results on MovieNet~\cite{huang2020movienet}~\cite{rao2020local} dataset in Figure~\ref{qua_mn}. This corresponds to Table 3 and Table 4 in the main paper and includes examples on place tagging as well as scene boundary detection (SBD) tasks. We show three examples from test set of MovieNet, and in each example, there are two scenes divided by the green dotted line. For each scene, the task is to predict what are the multi-label place tags of the scene, and for the two scenes together, the goal is to predict whether the shot boundaries between each pair of shots are also scene boundaries.

We can see that for SBD, our model can perform well to clearly identify the scene boundaries. For place tagging, it is a much more difficult task involving holistic understanding of the scenes, and although our model outperformed existing state-of-the-art models by a large margin in Table 3 in the main paper, it is still a really challenging and unsolved task. For example, some places are not easy to identify based on a few frames (e.g. playground), and some places can vary a lot in term of appearance but have same place tag (e.g. car). This is also partially caused by the lack of labeled data, where for the 90-category multi-label problem, there are $19.6\textrm{K}$ place tags, with $\sim$$11.7\textrm{K}$ for training, leading to $\sim$$130$ labeled training tags per category on average.

\subsection{MCD}

\setlength{\tabcolsep}{2pt}
\begin{table}[t]
	\begin{center}
		\smaller{
			\begin{tabu}{c|c|ccc|c}
				\hline
				Models & Pre-training data & $\mathsf{sex}$  &	$\mathsf{violence}$ & $\mathsf{drug}$-$\mathsf{use}$ & $\mathsf{average}$	\\\tabucline[1.0pt]{-}
				ShotCoL \cite{Chen_2021_CVPR}	&movie shot pairs	&62.3	&58.7	&47.1	&56.0\\
				MerlotReserve \cite{zellers2022merlotreserve}	&Youtube videos	&77.0	&68.2	&53.4	&66.2\\
				BridgeFormer \cite{ge2022bridgeformer}	&image+video	&74.8	&61.4	&61.2	&65.8\\
				Ours & MovieCL30K & \textbf{81.5} & \textbf{70.2} & \textbf{61.8} & \textbf{71.1}\\\hline
			\end{tabu}
		}
	\end{center}
	\vspace{-0.5cm}\caption{\small{Comparisons on $\mathsf{MCD}$ with other pre-trained models.}}
	\vspace{-0.3cm}
	\label{mcd}
\end{table}

\noindent Representative examples from the three age-appropriate activities in $\mathsf{MCD}$ dataset are provided in Figure~\ref{eg_cd}. We also show the samples for each of the 4 classes of our $\mathsf{MCD}$ dataset in Figure~\ref{qua_mcd} along with the corresponding detection results (i.e., the class with maximum probability) from both our model and CLIP model. For $\mathsf{sex}$ examples, we can see that our model has higher confidence scores compared to CLIP model especially when the images have dark illumination. For $\mathsf{violence}$, the CLIP model sometimes mistakes scenes with two closed persons as $\mathsf{sex}$ as shown in the first example. Similarly for $\mathsf{drug}$-$\mathsf{use}$ examples, our model classifies them more confidently, and CLIP model misses the small cigarette in the third example and classifies it as $\mathsf{none}$. The $\mathsf{none}$ examples show that the CLIP model often mistakes them with other classes, such as $\mathsf{violence}$ and $\mathsf{drug}$-$\mathsf{use}$, while our model is able to classify them correctly. This indicates our scene representation performs better than CLIP on age-appropriate activities, demonstrating the effectiveness of our representation in video moderation. Additional quantitative results on $\mathsf{MCD}$ are provided in Table~\ref{mcd} for comparisons with other pre-trained large models. 

\section{Additional Insights}

\noindent \textbf{a. Details of scene adjacency matrix:} Consider an example movie-pair $\mathbf{x_1}$ and $\mathbf{x_2}$ with $\mathbf{m}$ and $\mathbf{n}$ number of scenes respectively. The shape of their scene adjacency matrix $\mathbf{B}$ is $\mathbf{m}$$\times$$\mathbf{n}$. Each value in $\mathbf{B}$ indicates the similarity score between two scenes, one from each movie. We rank all the similarity scores in $\mathbf{B}$ so that we know which pairs of scenes are most similar. We first select the scene-pair (say $\mathbf{m}_{\textrm{0}}$ and $\mathbf{n}_{\textrm{0}}$) in $\mathbf{B}$ with the highest similarity score and add it to $\mathbf{P}_{\textrm{scene}}$. Moreover, we keep a record of this scene-pair, so that $\mathbf{m}_{\textrm{0}}$ and $\mathbf{n}_{\textrm{0}}$ will not be selected again from movie-pair $\mathbf{x_1}$ and $\mathbf{x_2}$. We then move on to the scene-pair corresponding to the second highest value in $\mathbf{B}$, and only add it to $\mathbf{P}_{\textrm{scene}}$ if neither of the scenes in that scene-pair is $\mathbf{m}_{\textrm{0}}$ or $\mathbf{n}_{\textrm{0}}$. We carry out this process for the top $50\%$ of the most similar scene-pairs in $\mathbf{B}$. This movie-pair level routine is repeated for all pairs of movies in our dataset to build $\mathbf{P}_{\textrm{scene}}$.

\noindent \textbf{b. Effectiveness of pre-trained CLIP weights:} In general, it has been demonstrated that pre-training does not always result in improvements \cite{He_2019_ICCV}. Specifically for our case, there are two key reasons why using pre-trained CLIP does not offer additional benefits.
%These were not pre-trained CLIP but the observations hold in our case. Also, pre-training generally helps more significant when the data used in fine-tuning is limited (e.g., a few hundred or thousand).
First, the domain gap between pre-trained CLIP (internet images and texts) and our data (movies) is quite high. %Meanwhile, we have $2.5$M samples for scene representation learning, and the feature space we are learning is diverging significantly from CLIP.
%Thus, even when we use pre-trained CLIP weights as initialization, over the course of training, the weights have been changed significantly. Otherwise, the results on row 14 in Table 1 of main paper will not significantly outperform the ones on row 6 in Table 2 of main paper.
Second, CLIP uses individual images and incorporates no temporal information. However, our use of scenes heavily relies on information among frames and shots.

%%%%%%%%% REFERENCES
{\small
\bibliographystyle{ieee_fullname}
\bibliography{ref_supp}
}